%% file: main.tex
\documentclass{article}

\usepackage[utf8]{inputenc}
\usepackage[T1]{fontenc}

\usepackage{arxiv}

\usepackage{url}
\usepackage{amsfonts}
\usepackage{nicefrac}
\usepackage{microtype}
\usepackage[normalem]{ulem}

\usepackage{graphicx}
\graphicspath{ {./images/} }

\usepackage{xcolor}
\usepackage{booktabs}
\usepackage{multirow}
\usepackage{colortbl}
\usepackage{array}

\usepackage{float}
\usepackage{caption}
\usepackage{subcaption}
\usepackage{placeins}

\usepackage{siunitx}
\usepackage{listings}

\usepackage{subfiles}

\usepackage[backend=biber, style=apa, uniquename=false]{biblatex}
\usepackage{hyperref}

\title{LLM-Driven Performance-Space Augmentation for Meta-learning-Based Algorithm Selection\thanks{Code is available at \url{https://github.com/lxt3rm/llm_synth_generator}.}}

\author{
 Darren Zhu \\
  Department of Statistics and Data Science\\
  National University of Singapore\\
  \texttt{e1121541@u.nus.edu} \\
  \And
  Daren Ler \\
  Department of Computer Science\\
  National University of Singapore\\
  \texttt{dcsdlsw@nus.edu.sg} \\
}

\lstdefinestyle{prompt}{
  basicstyle=\ttfamily\small,
  breaklines=true,
  breakatwhitespace=false,
  columns=fullflexible,
  keepspaces=true,
  frame=single,
  framesep=4pt,
  xleftmargin=6pt,
  xrightmargin=6pt,
  showstringspaces=false,
  upquote=true,
}

\definecolor{bestcol}{RGB}{198, 239, 206}
\definecolor{headerbg}{RGB}{68, 84, 106}
\definecolor{headerfg}{RGB}{255, 255, 255}
\definecolor{subheadbg}{RGB}{217, 225, 242}
\definecolor{posgreen}{RGB}{0, 140, 60}
\definecolor{posbrown}{RGB}{180, 100, 0}

\newcommand{\poscolor}[1]{\textcolor{posgreen}{\textbf{#1}}}
\newcommand{\posweak}[1]{\textcolor{posbrown}{\textbf{#1}}}
\newcommand{\best}[1]{\cellcolor{bestcol}\textbf{#1}}

\begin{document}
\maketitle

\begin{abstract}
Meta-learning for algorithm selection relies on a meta-dataset in which each row corresponds to a supervised learning dataset described by meta-features and labelled with a target value that is associated with algorithm choice (typically, some function of algorithm performance). A persistent limitation is that the number of curated real-world datasets is small, resulting in sparse meta-datasets that constrain meta-learner generalisation. In this paper, we address this problem by augmenting the meta-dataset with synthetic regression datasets produced via a large language model (LLM), with generation steered toward target regions of a low-dimensionality \emph{performance space}. In our experiments, we adopt a two-dimensional geometric setting defined by the cross-validated $R^2$ scores of two anchor algorithms, known as landmarkers. We compare two augmentation strategies: (1) uniform sampling, which distributes synthetic datasets across the performance space; and (2) margin-based sampling, which concentrates them near the decision boundary where landmarker preference is most ambiguous. Across 42 real-world UCI regression datasets and 730 synthetic datasets, both strategies substantially improve meta-learner performance over the unaugmented baseline under regression and multi-label evaluation formulations. However, uniform augmentation consistently outperforms margin-based augmentation, achieving a \num{17.47}\% relative reduction in Hamming loss, a \num{100.41}\% relative improvement in subset accuracy, and a \num{+6.09}\% relative gain in pooled out-of-fold $R^2$. These results lead us to postulate a central thesis. \textit{The performance of algorithms resides on a low-dimensional performance manifold, whose reconstruction bias may be minimised via the application of user-guided LLMs that seek to maximise uniform $\epsilon$-cover. The reduction of this reconstruction bias would lead to improved meta-learning for algorithm selection.}

\end{abstract}

\section{Introduction}

\subfile{sections/intro}

\section{Related Literature}

\subfile{sections/lit_review}

\section{Proposed Method for Synthetic Dataset Generation}

\subfile{sections/method}

\section{Experimental Setup}

\subfile{sections/experiment}

\section{Analysis of Results}

\subfile{sections/results}

\section{Conclusion}

\subfile{sections/conclusion}

\newpage
\printbibliography

\newpage
\section{Appendix}

\subfile{sections/appendix}

\end{document}

%% file: sections/intro.tex
Algorithm selection corresponds to the problem of choosing the most suitable algorithm for a given dataset. As no single algorithm is universally optimal over all possible datasets \parencite{wolpert1996nfl1, wolpert1996nfl2}, we are required to match an algorithm's inductive bias to the properties of the dataset. This makes algorithm selection a central concern in applied machine learning. Meta-learning provides a principled framework for this problem. By learning relationships between dataset characteristics (meta-features) and algorithm performance from a collection of prior datasets, a meta-learner can recommend algorithms for new, unseen datasets \parencite{rice1976algorithm, smith2009cross}.

As with most learning problems, the effectiveness meta-learning depends critically on the data, or in this case, the meta-dataset. The meta-dataset corresponds to a collection of previously observed datasets from which the meta-learner is trained. Each (base-level) dataset contributes one meta-example, so the size of the meta-dataset is bounded by the number of available datasets. 

In practice, curated dataset repositories such as UCI \parencite{uci_ml_repository} provide only a limited number of datasets for any given problem type. For regression algorithm selection, \textcite{lorena2018} worked with just 39 real-world regression datasets, and \textcite{soares2009uci++} noted that much empirical meta-knowledge was built from only around 100 datasets. This meta-dataset scarcity limits the generalisability of the meta-learner, particularly when the meta-feature space is high-dimensional \parencite{reif2012GA}.

Prior work has addressed this bottleneck through synthetic dataset generation \parencite{soares2009uci++, reif2012GA, komorniczak2025epco}. However, as these approaches operate within the high-dimensional meta-feature space, the problem of generating datasets remains challenging. Consequently, much of the work focuses on whether synthetic datasets \textit{can} be generated with certain properties, rather than on which \textit{augmentation strategy} best supports meta-learner generalisation.

%Prior work has addressed this bottleneck through synthetic dataset generation. \textcite{soares2009uci++} proposed datasetoids---new datasets derived by transforming existing ones---while \textcite{reif2012GA} developed a genetic algorithm that evolves datasets to match target meta-feature values. More recently, \textcite{komorniczak2025epco} extended this direction by generating datasets with controlled complexity levels. These approaches operate in meta-feature space, targeting specific meta-feature values or complexity measures. However, specifying a meaningful target distribution over a high-dimensional meta-feature space is difficult without strong assumptions about the deployment problem distribution. Moreover, existing work focuses primarily on whether synthetic datasets \textit{can} be generated with certain properties, rather than on which augmentation \textit{strategy} best supports meta-learner generalisation.

In parallel, large language models (LLMs) have emerged as capable generators of synthetic tabular data \parencite{zhang2026llm_generation_survey}, demonstrating that they can be guided toward producing data with desired structural properties. While this work focuses on sample-level augmentation within a fixed source dataset, it suggests that LLMs can serve as flexible generators when combined with appropriate constraints and evaluation mechanisms.

In this paper, we propose a different approach to \textbf{synthetic meta-dataset augmentation}. 

Firstly, rather than guiding generation in meta-feature space, we frame augmentation in \textit{performance space}, a low-dimensional space defined by the cross-validated $R^2$ scores of candidate algorithms. This framing aligns directly with the algorithm selection objective. Our empirical results expose a potentially exploitable \textit{structural embedding} within this space (i.e., a latent structure of comparative algorithm behaviour). 

Secondly, to probe the properties of the algorithm performance space, we compare two augmentation strategies: uniform- versus margin-based sampling. The former assumes a meta-learner requires global coverage of the performance space, whereas the latter assumes that only local refinement near the decision boundary is necessary. 

Thirdly, we use an LLM-based generator to produce synthetic regression datasets whose algorithmic behaviour (i.e., performance) falls within specified target regions of this performance space. 

%Within this framing, we first ask whether performance-space augmentation improves meta-learner generalisation. Conditional on a positive answer, we then ask which of two concrete sampling strategies better realises that improvement: (1) uniform sampling, which provides broad coverage across the performance space; or (2) margin-based sampling, which concentrates synthetic datasets near the decision boundary where algorithm preference is most ambiguous.

These strategies are evaluated on a meta-learning task for regression algorithm selection using 42 real-world UCI datasets and 730 synthetically generated datasets. Our experiments show that both strategies substantially improve meta-learner generalisation over the unaugmented baseline, and that uniform augmentation consistently outperforms margin-based augmentation across all evaluation metrics. 

These findings support the viability of performance-space-guided augmentation and suggest that broad coverage is more valuable than concentrated boundary sampling for the meta-learners studied.

The remainder of this paper is organised as follows. Section~2 reviews related work on meta-learning for algorithm selection, synthetic dataset generation, and LLM-based tabular data generation. Section~3 presents the proposed method, including the hypotheses, the LLM-based generator, and the two augmentation strategies. Section~4 describes the experimental setup and evaluation metrics. Section~5 presents the analysis, including Monte Carlo validation, meta-feature characterisation, and ablation and learning curve results. Section~6 concludes and discusses future directions.

%% file: sections/lit_review.tex
\subsection{Meta-Learning for algorithm selection}
The algorithm selection problem concerns choosing the algorithm most likely to perform best on a given dataset. Under the classical formulation by \textcite{rice1976algorithm} and again refined by \textcite{smith2009cross}, the algorithm selection problem is decomposed into four components: (i) a problem space $\mathcal{P}$, (ii) a feature space $\mathcal{F}$, (iii) an algorithm space $\mathcal{A}$, and (iv) a performance space $\mathcal{Y}$. Under this view, a dataset $x \in \mathcal{P}$ is described by a meta-feature vector $f(x) \in \mathcal{F}$, and the goal is to learn a selection mapping $S: \mathcal{F} \rightarrow \mathcal{A}$ such that the chosen algorithm $\alpha \in \mathcal{A}$ is expected to maximise performance $y(\alpha(x)) \in \mathcal{Y}$. This formulation is especially relevant in machine learning because the No Free Lunch theorems \parencite{wolpert1996nfl1, wolpert1996nfl2} imply that no single learner is uniformly optimal across all problems. Instead, performance depends on how well an algorithm’s inductive bias matches the structural properties of the dataset. Accordingly, meta-learning has emerged as a principled approach to algorithm selection by learning the mappings between problem characteristics and algorithm performance from prior experience.

Consequently, algorithm selection is treated as a meta-level supervised learning problem. A meta-dataset is constructed from previously observed problems, where each meta-example corresponds to a dataset or problem instance. The inputs are structural descriptors (meta-features) characterising that problem, while the target captures an aspect of algorithm performance or recommendation.

Useful taxonomies of meta-learning systems for classifier selection are provided by \textcite{khan2020survey} and \textcite{vanschoren2019survey}, which characterise prior work along three dimensions: meta-features, meta-learners, and meta-targets. Within these surveys, meta-features are defined as descriptors used to characterise datasets; meta-learners are the models used to map those descriptors to recommendations; and meta-targets specify the specific predictive objective at the meta-level. The surveys categorises meta-features into simple/statistical/information-theoretic \parencite{brazdil1994statlog, won2008simple}, complexity-based \parencite{ho2002complexity}, model-based \parencite{bensusan1998model-based}, landmarking-based \parencite{bensusan2000landmarking, pfahringer2000landmarking, ler2004landmarker}, and structural-information-based descriptors \parencite{tatti2007structural, song2012structural}. Meta-learners are categorised into rule-based \parencite{ali2006rule, brazdil1994rule}, regression-based \parencite{reif2014regression, garcia2016regression, bensusan2001regression}, instance-based \parencite{wang2013instance-based, zhang2019instance-based, brazdil2003instance-based}, multi-label-based \parencite{wang2014multi-label, wang2015multi-label}, and link-prediction-based methods \parencite{zhu2018link-based}. Meta-targets include best-algorithm prediction \parencite{garcia2016regression, reif2014regression, bensusan2001regression}, ranked-list recommendation \parencite{brazdil2003instance-based, song2012structural}, and multiple-algorithm recommendation \parencite{wang2015multi-label, wang2014multi-label}. 

However, much of this meta-learning literature has historically focused on classification, despite the general framework's applicability across domains. \textcite{smith2009cross} explicitly notes that meta-learning concepts were only limitedly generalised beyond classification in early research, creating a tension between the broad generality of Rice’s framework and the practical reality that informative dataset descriptors are often problem-specific. In other words, while the overall architecture of meta-learning is generic, its success depends heavily on whether the selected meta-features adequately characterise the underlying structure of the target problem paradigm.

This issue becomes clear in regression. As \textcite{lorena2018} observed, regression meta-learning has long relied on descriptors adapted from classification or on generic statistical summaries of the target variable, such as variance, skewness, and kurtosis. Earlier regression work also used landmarking \parencite{kuba2002landmarking-regression, soares2004landmarking-regression, amasyali2009landmarking-regression} and algorithm-specific measures such as kernel target alignment \parencite{gomes2012kta, soares2006kta, cristianini2001kta}, but did not formally characterise regression complexity in the same manner as classification. \textcite{lorena2018} argue that this represents a substantive gap, as regression problems possess unique sources of difficulty, including the linearity or nonlinearity of the input-output relation, the smoothness of the target function, the informativeness of features, and the sparsity of the input space. To address this, \textcite{lorena2018} proposed regression-specific complexity meta-features, demonstrating that these measures describe regression datasets more effectively and perform better than classical meta-features in empirical evaluations. 

Beyond the choice of meta-features, a recurrent practical limitation in meta-learning is that the number of available real-world datasets is often small relative to the dimensionality of the meta-feature space. Since each dataset contributes only a single meta-example, the resulting meta-dataset is frequently sparse, which in turn limits the generalisability of the meta-learner. This issue was highlighted in earlier work on meta-learning for algorithm selection, where \textcite{soares2009uci++} noted that much empirical meta-knowledge was historically derived from only approximately 100 datasets, making it difficult to construct robust meta-learners. \textcite{reif2012GA} made a similar observation, arguing that meta-feature space is often high-dimensional yet only sparsely populated, necessitating additional datasets for meta-learners to achieve greater predictive power.

Taken together, the literature establishes a clear foundation for meta-learning for algorithm selection while underscoring an unresolved meta-dataset size bottleneck. For example, while \textcite{lorena2018} demonstrate that regression-specific complexity descriptors are informative, their study was limited to 39 real-world regression datasets, resulting in a meta-dataset of only 39 instances. This gap directly motivates our work. If meta-learning for regression algorithm selection depends critically on both the quality of descriptors and the geometric coverage of the meta-dataset, then generating additional synthetic regression datasets represents a promising strategy for improving meta-learner generalisation. Next, we examine previous approaches to this topic.

\subsection{Synthetic data generation for augmenting the meta-dataset}

\textcite{soares2009uci++} addresses the meta-dataset scarcity problem by deriving new datasets from existing ones. He introduces the concept of "datasetoids," generated by a transformation in which a categorical feature is swapped with the target variable, thereby producing additional datasets from existing real-world data. Using 64 UCI datasets, this procedure yielded 983 datasetoids, substantially enlarging the pool of meta-examples available for training. In a meta-learning task for predicting whether decision-tree pruning would be beneficial, the inclusion of datasetoids improved meta-level accuracy for several learners relative to using only the original UCI datasets, suggesting that transformed datasets can supply useful additional meta-knowledge. At the same time, the paper notes that datasetoids may exhibit anomalies and may not correspond to meaningful real-world problems; thus, their utility depends on whether they remain representative of the target problem space $\mathcal{P}$.

\textcite{reif2012GA} propose another solution involving the direct generation of datasets with specified meta-feature values. They introduce a genetic data generator specifically designed for meta-learning research. Their core idea is to treat data generation as an optimisation problem, where candidate datasets are evolved so their measured meta-features match user-specified target values. In this formulation, each individual in the genetic algorithm is a complete dataset defined by its data points; the initial population is created by sampling feature values from normal or uniform distributions with random parameters. The fitness function is a weighted sum of absolute deviations between the measured meta-features and their desired target values:
$$f(x) = \sum_{i=1}^{n} w_i \cdot |x_i - y_i|$$
where $x$ is the meta-feature vector of a candidate dataset, $y$ is the vector of desired values, and $w$ is the corresponding weight vector. Because the optimal value of this objective is zero, the algorithm terminates once fitness falls below a specified threshold. The search process employs Gaussian mutation to shift individual data points and two-point crossover to swap fractions of data between instances, with the implementation utilising the DEAP evolutionary computation framework.

This approach allows synthetic datasets to be placed more deliberately in under-represented parts of the meta-feature space to improve coverage. The authors argue that such a generator supports meta-learning by filling "empty areas" of a sparse meta-feature space in a controlled manner. Importantly, while flexible, their implementation was limited to numerical features and classification datasets, leaving a gap for more complex problem classes such as regression.

A more recent direction moves from matching generic meta-features toward controlling problem complexity more explicitly. \textcite{komorniczak2025epco} proposes a genetic algorithm, called Evolutionary Projection-based Complexity Optimisation (EPCO), that transforms synthetically generated source datasets so that selected complexity measures approach desired target values. Unlike \textcite{reif2012GA}, where individuals represent raw datasets, EPCO represents each individual as a $d \times d$ transformation matrix, where $d$ is the feature dimensionality. Matrix coefficients are initialised by sampling from $\mathcal{N}(0, 3)$. A candidate dataset is obtained by multiplying the original feature matrix by an individual's transformation matrix, leaving the original labels unchanged. For each complexity measure $C_i$, fitness is defined as $F_i = |T_i - C_i|$, the absolute difference between the target value $T_i$ and the measured value, resulting in a many-objective minimisation problem. The population is ordered so that the leading individuals demonstrate superior fitness across each criterion and in their aggregate sum. Crossover combines a leading individual with a randomly selected one via a weighted sum of transformation matrices, while mutation introduces small random noise $\mathcal{N}(0, 0.1)$ to a randomly selected individual. Crossover and mutation ratios are reduced over iterations by a decay factor to stabilise the search in later generations. 

EPCO is developed for both classification and regression; for classification, ten complexity measures are optimised, while for regression, four measures from \textcite{lorena2018} with promising optimisation behaviour were selected. Experiments indicate that datasets with varying target difficulty levels can be produced, and that these complexity levels correlate with downstream recognition quality. This is especially relevant for regression meta-learning as it suggests that synthetic augmentation need not merely increase the quantity of meta-examples, but should be designed to expand coverage across levels of problem difficulty that directly influence algorithm behaviour.

Collectively, these studies suggest a progression toward more structured approaches to augmenting the meta-dataset. The first involves dataset transformation, where additional meta-examples are obtained efficiently from existing real-world datasets, as seen in datasetoids. The second is meta-feature-targeted generation, where synthetic datasets are evolved to occupy specific regions of the feature space $\mathcal{F}$. Building upon these is complexity-targeted generation, where the augmentation process is guided by measures reflecting learning difficulty more directly. Across all three approaches, the shared motivation is not merely to increase data volume, but to improve the coverage of the meta-dataset, enabling the meta-learner to infer a more reliable mapping from dataset characteristics to algorithm performance.

However, while prior work establishes that augmenting the meta-dataset is feasible and often beneficial, it leaves open the more targeted question of which specific augmentation policy produces the most informative meta-examples for algorithm selection. This gap defines the positioning of our study. Compared to earlier work, our setting is more specialised in two aspects. First, we focus specifically on meta-learning for candidate regression algorithm selection, where the utility of augmentation depends on how effectively synthetic datasets help the meta-learner model variations in regressor performance. Second, rather than assessing only the feasibility of generating datasets with certain properties, we compare two distinct augmentation policies: (1) uniform augmentation, and (2) margin-based augmentation, to determine which better enhances meta-learner generalisation. In this sense, our work extends the literature from the mechanics of dataset generation to the decision-oriented question of how synthetic datasets should be sampled or targeted to optimise the meta-learning process.

\subsection{LLM for synthetic tabular generation}
Recent research has demonstrated that Large Language Models (LLMs) can be used effectively for synthetic tabular data generation, particularly in settings where the objective is to generate realistic samples for an existing dataset \parencite{zhang2026llm_generation_survey}. In this work, the central challenge is to model heterogeneous numerical and categorical attributes jointly while preserving cross-column dependencies, schema-level validity, and statistical fidelity to the source distribution. Existing methods can be broadly grouped into prompt-based approaches \parencite{kim2024prompt, fang2025prompt, seedat2023prompt}, fine-tuned tabular generators \parencite{borisov2022finetune}, and hybrid structured systems \parencite{zhang2025hybrid}. 

Prompt-based methods rely on instruction design, in-context exemplars, and conditional prompting to improve realism and long-tail coverage, whereas fine-tuned approaches, such as GReaT and its extensions, train autoregressive models directly on serialised samples to improve structural consistency. Hybrid methods further combine LLM reasoning with external components for constraint handling and statistical alignment. Collectively, these studies show that LLMs are promising generators of tabular data, especially when augmented with mechanisms that enforce validity and dependency constraints.

Nevertheless, the dominant formulation in this literature remains instance-level augmentation (i.e., within dataset instance generation), where the aim is typically to generate additional samples that resemble those from a specific real-world dataset. 

Our setting differs in a more fundamental way. Rather than augmenting a fixed table, we seek to generate entire synthetic datasets for meta-learning, with the goal of inducing controlled variation in downstream algorithm performance. Thus, whereas prior tabular synthesis methods prioritise fidelity to a single source distribution, our problem requires dataset-level generation across multiple target behaviours. Existing LLM-based tabular generation work is therefore most useful to us as conceptual inspiration; it demonstrates that LLMs can be guided toward desired tabular properties, but it does not directly address the problem of generating diverse whole datasets whose meta-level utility lies in how they reshape the performance landscape for algorithm selection.

%% file: sections/method.tex
The proposed method is motivated by the view that the effectiveness of meta-learning for algorithm selection depends not only on the choice of meta-features or the meta-learner architecture, but also on the breadth and structural density of the meta-dataset used for training. Because the number of curated real-world regression datasets is limited (e.g. from the UCI repository), they often fail to provide sufficient coverage for complex meta-learners, and the resulting meta-dataset may fail to adequately represent the full range of problem regimes relevant to algorithm selection. This motivates the use of synthetic datasets as a controlled means of expanding the meta-dataset to cover these under-represented regions. 

In this work, we propose that \textbf{the generation process be framed in performance space} $\mathcal{Y}$ \textbf{rather than the full meta-feature space} $\mathcal{F}$. This shift provides a more interpretable and operationally tractable basis for defining coverage relative to the algorithm selection objective. Interpretability arises because performance-space coordinates directly encode comparative algorithm behaviour, where each axis represents the performance metric of a candidate algorithm on a given dataset, making augmentation targets immediately meaningful for the algorithm selection objective. Operational tractability follows from the lower dimensionality of the performance space; for instance, specifying a target region in a two-dimensional performance plane is significantly simpler than prescribing a joint distribution over a high-dimensional meta-feature space whose deployment-relevant structure is generally unknown.

\subsection{Hypotheses}\label{sec:hypotheses}

We hypothesise that augmenting a meta-dataset, comprised of real-world regression datasets, with synthetic regression datasets, can improve the generalisation of a meta-learner for algorithm selection, provided that the synthetic datasets are generated in a principled manner. The underlying premise is that the predictive quality of a meta-learner is constrained not only by the learning algorithm used at the meta-level, but also by the \textit{structural coverage} of the problem space induced by the training meta-dataset. When that meta-dataset is constructed solely from a limited pool of curated real-world datasets, important problem regimes may be under-represented, reducing the meta-learner's ability to generalise to previously unseen datasets. From this perspective, synthetic dataset generation is valuable not merely for increasing the number of meta-level training instances (i.e., base-level datasets), but as a mechanism for improving coverage of informative regions within the problem space.

While previous works focus on generating synthetic datasets in the meta-feature space $\mathcal{F}$ \parencite{soares2009uci++, reif2012GA, komorniczak2025epco}, we instead formulate this coverage in the performance space $\mathcal{Y}$. 

Essentially, this is rationalised from an observation made in  \textcite{ler2004landmarker}, which introduced the performance space as a setting for landmarker-based algorithm selection. In that work, it was proposed that correlation- and efficiency-based criteria could be used to identify a small set of informative landmarker axes that carve out this space. This supports the argument that algorithm ``domains of expertise'' inhabit a lower-dimensional projection than contemporary high-dimensional meta-features suggest.

Consequently, we postulate that the specification of a meaningful target distribution over high-dimensional meta-features will be difficult without strong assumptions about the deployment problem distribution. In contrast, the performance space is directly aligned with the algorithm selection objective, as it represents comparative algorithm behaviour rather than abstract dataset descriptors alone. Furthermore, when landmarking information is included among the meta-features, the performance space may be interpreted as a partial projection of the broader meta-feature representation, making performance-space targeting a practically meaningful way to structure synthetic dataset generation.

Accordingly, the proposed method is guided by the following hypotheses.

\textbf{H1: Synthetic-augmentation hypothesis}. Augmenting a meta-dataset that is comprised of real-world problem datasets, henceforth denoted the \textit{real-world meta-dataset}, with synthetic regression datasets improves the out-of-sample generalisation of the meta-learner. The null hypothesis is that augmenting the real-world meta-dataset with synthetic regression datasets does not improve out-of-sample meta-learner performance over training on the real-world meta-dataset alone. The alternate hypothesis is that augmentation does improve out-of-sample meta-learner performance.

\textbf{H2: Sampling-strategy hypothesis} (conditioned on H1). Within performance-space-targeted augmentation, the choice of sampling strategy significantly affects meta-learner performance. The null hypothesis is that uniform and margin-based sampling produce equivalent out-of-sample meta-learner performance. The alternate hypothesis is that they differ. In the two-dimensional performance space defined by the cross-validated $R^2$ scores of the two anchor algorithms, the \textit{decision boundary} is the diagonal $y = x$ where both algorithms perform equally. Margin-based sampling concentrates synthetic datasets near this boundary to refine the decision edge, while uniform sampling distributes them across the full performance space to maintain global cover.

Taken together, H1 and H2 define the design rationale of the proposed method. H1 addresses whether synthetic dataset augmentation is beneficial, while H2 asks how synthetic datasets should be allocated across the performance space to best support meta-learning. Specifically, a consistent advantage of uniform sampling over margin-based sampling would indicate that the meta-learner requires a global cover of the performance space rather than localised refinement at the decision boundary.

\subsection{Proposed LLM Synthetic Dataset Generator}\label{sec:llm_generator}

An unguided generator, which replies on random sampling over candidate $(X, y)$ pairs, satisfies the $\varepsilon$-cover only in the limit and at intractable computational cost. We define an $\epsilon$-cover as a structural condition where every point in the target region of the performance space $\mathcal{Y}$ has at least one synthetic dataset "witness" within a distance $\epsilon$, effectively ensuring no "blind spots" exist in the training data.

To operationalise this, we define a mapping function $\phi(D)$ that transforms a raw dataset $D$ into its \textit{structural embedding} coordinates. In our implementation, $\phi(D) = (R^2_{KNN}, R^2_{LR}) \in [0, 1]^2$, representing the mean cross-validated performance scores of K-Nearest Neighbours (KNN) and Linear Regression (based on ordinary least squares) (LR). This two-dimensional space is discretised into a rectangular grid where the unit square is partitioned into several bins per axis. Synthetic augmentation is thus cast as a task of populating selected grid cells with accepted datasets, ensuring coverage is guided by regions of comparative algorithm behaviour rather than raw dataset descriptors.

The generator addresses the challenge of achieving the $\epsilon$-cover through a \textit{propose--execute--evaluate--repair} loop. We formulate this as a constrained program-synthesis problem, i.e., rather than asking the LLM to produce raw data, the system prompts the model to return an executable Python procedure of the form \texttt{generate(seed)} $\rightarrow$ \texttt{(X, y)}. The prompt provides the specific target intervals for the current grid cell and explicitly encourages the model to exploit known differences between local and global regression behaviours, such as non-linearity or noise, to land the dataset within the requested $\phi$-target. 

\textbf{Execution and Validation}. Each proposed generator is executed in a spawned process under a restricted runtime to prevent unauthorised file-system access. The sandbox permits ordinary Python control flow and \texttt{NumPy} but blocks unrestricted imports. A critical feature of this environment is that it exposes an \texttt{evaluate(X, y)} function to the generated program. This allows the LLM to implement a \textit{lightweight within-procedure search}, which can scan over hyper-parameters (e.g., noise magnitude or dimensionality) to return the candidate dataset whose measured performance is closest to the requested cell (i.e., $\phi$-target).

\textbf{The Repair Loop}. Valid datasets are evaluated under a fixed comparison harness using mean $R^2$ over a deterministic, stratified five-fold splitter. If the measured coordinates fall outside the target cell, the system issues a \textbf{repair prompt} reporting the achieved $\phi$-coordinates. This request is sent as a continuation of the previous conversational thread to preserve context while asking the model to revise the mechanism. To isolate architectural changes from stochastic variation, all retries for the same cell reuse the same deterministically derived execution seed. For each target cell, the system attempts to collect $w = 10$ datasets (i.e., witnesses), subject to a total budget of $b = 84$ attempts.

The accepted dataset for a slot is therefore the output of a controlled problem-synthesis module designed specifically for testing hypotheses H1 and H2.

\subsection{Meta-Dataset Generation}\label{sec:meta_dataset_generation}

The meta-dataset is constructed at the problem level, with one row per regression dataset and columns comprising dataset-level descriptors (meta-features) together with meta-labels derived from the measured performance of a fixed candidate set of regressors. The unified pipeline stores both continuous (i.e., regression meta-label) and binary meta-label (i.e., multi-label meta-label) formulations for the same problem, allowing the resulting representation to be reused across different algorithm-selection settings. (Both are clarified later in this section.)

\textbf{Meta-feature generation}. For meta-feature extraction, each dataset is first preprocessed. Categorical features are one-hot encoded, missing values are removed, and all features are normalised. The meta-features are then computed using the regression module of \texttt{problexity} \parencite{komorniczak2023problexity}, which implements the 12 complexity measures (i.e., meta-features) proposed by \textcite{lorena2018}. These 12 measures characterise the dataset structure across several dimensions:

\begin{itemize}
    \item \textbf{Feature Efficiency} ($C1$–$C4$): Quantifies individual and collective feature-target correspondence through Spearman correlation and the fraction of difficult cases.
    \item \textbf{Linearity} ($L1$–$L3$): Measures residuals of a linear regressor and non-linearity on interpolated samples.
    \item \textbf{Local Smoothness} ($S1$–$S4$): Characterised through minimum-spanning tree output variation, input variation among neighbouring target values, leave-one-out 1-nearest-neighbour error, and 1-nearest-neighbour non-linearity on interpolated samples.
    \item \textbf{Dimensionality} ($T2$): The sample-to-dimension ratio.
\end{itemize}

Except for $T2$, all measures are computed after min-max normalisation of inputs and targets. Each value is converted to a scalar, with negative outputs clipped to zero, forming a final 12-dimensional meta-feature vector $f(x) \in \mathcal{F}$ for each dataset.

\textbf{Regression meta-label generation}. For the regression formulation, the meta-label associated with an algorithm family $a$ is a continuous estimate of its predictive performance on dataset $D$. The candidate set comprises five \texttt{scikit-learn} \parencite{scikit-learn} regressors: K-Nearest Neighbours, ordinary least squares linear regression, Lasso, Ridge, and Elastic Net. Each model is evaluated within a pipeline consisting of a \texttt{MinMaxScaler} followed by the regressor. 

Hyperparameter optimisation is performed separately for each dataset and each tunable algorithm using \texttt{Optuna} \parencite{akiba2019optuna} with a Tree-structured Parzen Estimator (TPE) sampler and a median pruner. The defined search spaces are:

\begin{itemize}
    \item \textbf{K-Nearest Neighbours}: $n\_neighbors \in \{1, \dots, \lfloor n/3 \rfloor\}$ and $weights \in \{uniform, distance\}$.
    \item \textbf{Ridge and Lasso}: $\alpha \sim \log \mathcal{U}(10^{-5}, 10^{1})$.
    \item \textbf{Elastic Net}: $\alpha \sim \log \mathcal{U}(10^{-5}, 10^{1})$ and $l1\_ratio \sim \mathcal{U}(10^{-3}, 1 - 10^{-3})$.
    \item \textbf{Linear regression}: Left untuned in its default configuration.
\end{itemize}

The tuning objective is the mean $R^2$ over an inner repeated regression-stratified cross-validation procedure. We implement the \textit{Totally Stratified Cross-Validation} (TSCV) routine \parencite{saez2022stratified}, which sorts the target values and assigns observations one by one to the currently smallest fold with random tie-breaking, thereby encouraging similar target distributions across the 10-fold inner stage.

Following tuning, the selected hyperparameter configuration is held fixed and re-evaluated under a separate outer TSCV protocol with 10 folds repeated 10 times. This yields 100 split-level $R^2$ values per algorithm. The continuous meta-label for algorithm $a$ is defined as the mean of these outer-split scores:
\[
y_{\mathrm{reg}}^{(a)}(D)=\frac{1}{|\mathcal{S}_{\mathrm{out}}|}\sum_{s\in\mathcal{S}_{\mathrm{out}}} R^2_s(a).
\]

Each dataset is therefore associated with a five-dimensional continuous performance vector. Because tuning is performed once at the dataset level and the resulting configuration is then reused across all outer splits, the implemented protocol is a \textit{two-stage tune--then--evaluate} procedure rather than fold-wise nested hyperparameter optimisation. This ensures that the final meta-labels reflect performance after model selection has occurred at the base level.

\textbf{Multi-label meta-label generation}. The multi-label formulation is derived from the same outer-split performance matrix used for regression. In this matrix, rows correspond to the 100 repeated outer TSCV splits and columns correspond to the five candidate algorithms.

For each split, algorithms are ranked in descending order of $R^{2}$, with average ranks assigned in the case of ties. To determine the meta-labels, we apply a two-step statistical procedure:

\begin{enumerate}
    \item \textbf{Friedman Test}: A Friedman chi-square test is conducted at a significance level of $\alpha = 0.05$ to determine if any statistically significant differences exist across the algorithm pool. If the null hypothesis is not rejected, all candidate algorithms are marked as applicable, yielding an all-ones label vector.
    \item \textbf{Nemenyi Post-hoc Test}: If significant differences are found, we calculate the Nemenyi \textit{Critical Distance} (CD):
        \[
        \mathrm{CD}=q_{\alpha}\sqrt{\frac{k(k+1)}{6N}},
        \]
    where $k$ is the number of algorithms (5), $N$ is the number of outer splits (100), and $q_{\alpha}$ is the studentised range statistic.
\end{enumerate}

Let $r^{*}$ denote the best (lowest) average rank observed for a dataset $D$. An algorithm $a$ receives a binary meta-label $y_{\mathrm{ml}}^{(a)}(D) = 1$ if its average rank $r_{a}$ satisfies:
$$r_{a} \leq r^{*} + \mathrm{CD}$$
Otherwise, the label is $0$. The resulting multi-label vector identifies the subset of algorithms whose outer-CV performance is statistically indistinguishable from the top-ranked method on that dataset, providing a robust target for the multi-label meta-learner.

\subsection{Synthetic Augmentation Strategy}\label{sec:augmentation_strategy}

Once the accepted synthetic regression datasets have been generated, each dataset is mapped into the same meta-representation as the real-world datasets, using the meta-features $f(x) \in \mathcal{F}$ and labels $y \in \mathcal{Y}$ defined previously, so that augmentation is performed within the meta-dataset space. Let $(X_{real}, Y_{real})$ denote the real-wrold meta-dataset and $(X_{syn}, Y_{syn})$ the synthetic pool.

During downstream evaluation, augmentation is applied fold-wise, i.e., for each training split of the real-world meta-dataset, $n$ synthetic rows are sampled and appended to the real-world training fold, while the test fold remains composed exclusively of real-world meta-instances. This preserves a clean real-world dataset evaluation protocol and prevents synthetic data from contaminating the held-out assessment. This augmentation procedure directly operationalises H1, while the following two strategies implement H2:

\begin{itemize}
    \item \textbf{Uniform Strategy} (\textbf{Global-Cover Sampler}): The $n$ synthetic rows are sampled uniformly without replacement from the available synthetic pool. Every accepted synthetic dataset has an equal probability of selection, irrespective of its location in performance space. This strategy treats augmentation primarily as a mechanism for overall coverage expansion.
    \item \textbf{Margin-Based Strategy} (\textbf{Boundary-Concentrated Sampler}): Sampling is biased toward synthetic datasets lying close to the decision boundary in the two-dimensional performance space. Boundary proximity for a dataset $j$ with coordinates $(s_{x,j}, s_{y,j})$ is measured by the perpendicular distance to the diagonal $y = x$:
    $$d_{j}=\frac{\left|s_{x, j}-s_{y, j}\right|}{\sqrt{2}}$$
    Sampling probabilities are obtained from a temperature-scaled softmax over these distances, such that datasets with smaller $d_j$ receive larger selection probabilities, and $n$ datasets are sampled without replacement from this induced distribution. This strategy prioritises datasets near the selection boundary where algorithm preference is most ambiguous and potentially most informative for the meta-learner.
\end{itemize}

%% file: sections/experiment.tex
\subsection{Experiment Setup}

We evaluate synthetic meta-dataset augmentation under two downstream meta-learning formulations defined on the same problem-level representation. In our experiment, the real-world meta-dataset contains 42 real-world datasets taken from the UCI repository, with each of these datasets represented by the 12 meta-features proposed by \textcite{lorena2018}, or more specifically, $C1$--$C4$, $L1$--$L3$, $S1$--$S4$, and $T2$. The same meta-feature vector matrix is used in both formulations. The synthetic augmentation pool is a separate meta-dataset in the same representation space. 

In our experiment, we followed the method described in Section~\ref{sec:llm_generator} and generated two sets of synthetic datasets in the two-dimensional performance space $(R^2_{KNN}, R^2_{LR})$. One set from a 5x5 grid which yielded 248 synthetic datasets, and a second set from a 7x7 grid which yielded 482 synthetic datasets. The 7x7 grid represents a higher-resolution sampling pass over the same performance space, providing a denser coverage. The combined synthetic pool contains 730 synthetic datasets. Please refer to the Appendix for a detailed analysis of the LLM generated synthetic datasets and its methodology.

The first formulation is a regression-based meta-learner that predicts the held-out performance of each candidate algorithm family. Here, we follow most of the evaluation set-up in \textcite{lorena2018}. Concretely, our evaluation script trains five separate regressors, one for each target column $ \{ R^2_{LR}, R^2_{KNN}, R^2_{\mathrm{lasso}}, R^2_{\mathrm{ridge}}, R^2_{\mathrm{elasticnet}} \} $. Each regressor is implemented as a \texttt{TransformedTargetRegressor} \parencite{scikit-learn} whose base estimator is a pipeline consisting of feature standardisation followed by a default Support Vector Regressor (\texttt{SVR}); the scalar target is also standardised inside each training fold (for both real-world and synthetic datasets). The choice of SVR as the meta-learner follows \textcite{lorena2018}, whose evaluation protocol we adopt for comparability. Since our purpose is to compare augmentation strategies, no model selection or hyperparameter tuning is performed for this meta-learner.

The second formulation is a multi-label meta-learner \parencite{wang2014multi-label, wang2015multi-label} that predicts the precomputed binary meta-label vector stored in the columns $\{\texttt{meta\_lr}, \texttt{meta\_knn}, \texttt{meta\_lasso}, \texttt{meta\_ridge}, \texttt{meta\_elasticnet}\}$. Multi-label classification is practically useful for algorithm selection because it allows for the recommendation of multiple suitable algorithms for a given dataset, rather than committing to a single best choice. This learner is implemented as binary relevance via \texttt{MultiOutputClassifier(GaussianNB())}, again preceded by feature standardisation. In this formulation, binary relevance means that one independent classifier is trained per algorithm label, with no modelling of dependencies between labels. Hence, one independent Gaussian Naive Bayes classifier is trained per label, and similar to the regression set-up, no classifier-chain dependency, calibration, or threshold tuning is introduced.

Both formulations use the same repeated cross-validation protocol. The scripts perform shuffled $K$-fold cross-validation with $K=10$ folds and 10 repeats by default. For repeat $r$, the real-world dataset partitioning is generated by \texttt{KFold(n\_splits=10, shuffle=True, random\_state=seed+r)}. Our experiment also uses a seed sweep for robustness, over 10 different seeds. For a fixed seed, the real-world dataset folds are identical across no augmentation, uniform augmentation, and margin-based augmentation, so differences between conditions arise only from the training data augmentation and not from different held-out partitions. Together, the shared-fold cross-validation and 10-seed sweep are what support H1's endpoint test (no-augmentation vs full-pool augmentation) and H2's strategy comparison (uniform vs margin-based augmentation) under matched conditions.

Augmentation is applied fold-wise and strictly to the training split. In each fold, the real-world dataset training fold is optionally augmented with $n_{\mathrm{syn}}$ sampled synthetic datasets, whereas the test fold remains composed exclusively of real-world datasets. Thus, synthetic datasets never participate in split formation and never contaminate held-out evaluation.

Under uniform augmentation, the script samples synthetic rows uniformly at random from the synthetic pool. 

Under margin-based augmentation, it first computes, for each synthetic dataset $j$, the distance
\[
d_j = \frac{|R^2_{KNN, j} - R^2_{LR, j}|}{\sqrt{2}},
\]
and then defines sampling probabilities by a temperature-scaled softmax, $p_j \propto \exp(-\alpha d_j)$, with $\alpha=10$ in the default implementation, which helps to create a "sharp" focus on the boundary $y=x$. These probabilities are computed once per evaluation run and reused across folds and repeats. The sampled synthetic rows are appended only to the training fold. In our experiments, sampling is done without replacement.

\subsection{Evaluation Metrics}

For the regression formulation, the evaluation script computes a pooled out-of-fold $R^2$ separately for each target algorithm. We adopt $R^2$ as the primary regression meta-learner metric following \textcite{lorena2018}, who also used it to evaluate meta-level regression performance. Let $y_{ik}$ denote the true held-out performance of algorithm $k$ on real-world dataset $i$, and let $\hat y_{ik}$ denote the corresponding out-of-fold prediction obtained by pooling predictions across all folds within one repeat. We report
\[
R^2_{\mathrm{OOF},k}
=
1 - \frac{\sum_i (y_{ik} - \hat y_{ik})^2}{\sum_i (y_{ik} - \bar y_k)^2},
\]
where $\bar y_k$ is the mean of the true target values over all real-world datasets. This is therefore a pooled OOF $R^2$, not an average of fold-wise $R^2$ values. We report these scores separately for each of the five targets. When a single scalar is needed for learning-curve analysis, we average the five target-wise seed means to obtain \texttt{mean\_target\_pooled\_oof\_r2}. %No ranking metrics, winner-selection accuracy, or recommendation-quality metrics are derived from the predicted performance vectors in the current experiment.

For the multi-label formulation, we first pool hard out-of-fold label predictions over all real-world datasets within a repeat, and then compute two overall multi-label metrics: subset accuracy and Hamming loss. Let $\mathbf{y}_i \in \{0,1\}^K$ denote the ground-truth binary label vector for real-world dataset $i$ and $\hat{\mathbf{y}}_i \in \{0,1\}^K$ the corresponding predicted label vector, where $K=5$ is the number of candidate algorithms and $N$ is the number of real-world datasets.

Subset accuracy measures the fraction of datasets for which the predicted label set exactly matches the ground truth:
\begin{equation}
\mathrm{SubsetAccuracy}
= \frac{1}{N}\sum_{i=1}^{N} \mathbf{1}\!\left\{\hat{\mathbf{y}}_{i} = \mathbf{y}_{i}\right\},
\label{eq:subset_accuracy}
\end{equation}
where $\mathbf{1}\{\cdot\}$ denotes the indicator function, which equals $1$ if its argument is true and $0$ otherwise.  A prediction is counted as correct only if all $K$ binary labels are simultaneously correct, making this a stringent exact-match metric over the full algorithm recommendation set.

Hamming loss measures the average fraction of individual algorithm labels that are incorrectly predicted:
\begin{equation}
\mathrm{HammingLoss}
= \frac{1}{NK}\sum_{i=1}^{N}\sum_{k=1}^{K}
\mathbf{1}\!\left\{\hat{y}_{ik} \neq y_{ik}\right\},
\label{eq:hamming_loss}
\end{equation}
where $y_{ik}$ and $\hat{y}_{ik}$ denote the ground-truth and predicted values of the $k$-th label for dataset $i$, respectively. Hamming loss is the primary multi-label metric used in our evaluation.
Hamming loss is more appropriate here because it evaluates each algorithm label independently, providing a granular measure of per-label prediction quality that is not dominated by the exact-match stringency of subset accuracy.

In both formulations, repeat-level scores are summarised by their mean and standard deviation across the 10 repeats for each seed. In seed-sweep mode, the final summaries are computed over seed-level means. We report the mean over seeds, the standard deviation over seeds, and the mean within-seed standard deviation. The learning-curve plots use these across-seed summaries and display mean $\pm$ one standard deviation bands.

\subsection{Comparison Protocol}

The comparison is therefore cleanly controlled. The no-augmentation baseline, uniform augmentation, and margin-based augmentation are all evaluated on the same real-world dataset folds for a fixed seed, and synthetic datasets only affect training. For the regression formulation, an improvement corresponds to a higher pooled OOF $R^2$ for a target, or a higher mean target pooled OOF $R^2$ when a single scalar summary is required. For the multi-label formulation, the primary notion of improvement is lower Hamming loss, and a higher subset accuracy.

Our experiment reports both fixed-budget and learning-curve summaries. Fixed-budget runs provide direct three-way comparisons at a chosen $n_{\mathrm{syn}}$ (e.g., $n_{\mathrm{syn}}=400$ in the saved runs), while the learning-curve driver repeats the same evaluation across multiple augmentation budgets and records, for each $n_{\mathrm{syn}}$, the best-performing strategy according to the primary metric direction. This makes it possible to study not only whether synthetic augmentation helps, but also how performance changes as the number of appended synthetic datasets increases. Specifically, the learning-curve sweep is used to test H1 at its endpoints (no-augmentation vs full-pool augmentation) and to test H2 across the full $n_{\mathrm{syn}}$ range.

To determine statistical significance, we run a two-sided paired Student's $t$-test (\texttt{scipy.stats.ttest\_rel}) on per-seed scores at $n=10$ seeds, with significance threshold $\alpha = 0.05$. Pairing across seeds is valid because the shared-fold property described above isolates the augmentation effect at each seed. For H1, the test is computed once per metric at the learning-curve endpoint ($n_{\mathrm{syn}} = 730$). Uniform and margin both sample the full synthetic pool there and collapse to a single augmentation arm, yielding three $p$-values. For H2, the test is computed at every train-size point of the learning curve, giving $3 \times 37 = 111$ (metric, train-size) cells. We summarise these as per-metric $p$-value-vs-train-size plots in the main text and tabulate them in full in the appendix. We report uncorrected $p$-values throughout.

Beyond the H1 and H2 comparisons above, we run a coverage-granularity comparison between the two grids in isolation. The 5x5 grid contributes \num{248} synthetic datasets, and the 7x7 grid contributes \num{482}. Each grid is used in turn as the full augmentation pool under the same 10x10 repeated cross-validation protocol and the same meta-learners as the main experiment. For each grid, we record per-seed scores across the same \num{10} seeds on the three metrics Pooled OOF $R^{2}$, Hamming Loss, and Subset Accuracy. Statistical significance is then assessed with the same two-sided paired Student's $t$-test (\texttt{scipy.stats.ttest\_rel}) at $\alpha = \num{0.05}$, computed once per metric on the seed-paired difference 7x7 minus 5x5. This yields three $p$-values, alongside mean paired differences and 95\% confidence intervals on the differences.

%% file: sections/results.tex
\subsection{Monte Carlo Analysis of Augmentation Sampling Strategies}
\label{sec:monte_carlo_analysis}

To characterise the behavioural difference between the uniform and margin-based augmentation strategies under H2,
a Monte Carlo frequency analysis was conducted over the synthetic candidate pool.

From the full pool of \num{730} synthetic meta-instances, $n_{\mathrm{syn}} = \num{400}$ instances (i.e., approximately 50\% of all instances, rounded to the nearest hundred)
were drawn without replacement in each iteration, and this process was repeated for
$N = \num{1000}$ independent iterations under both strategies.
The margin-based strategy assigns sampling probabilities via a temperature-scaled softmax over
the negative perpendicular distance of each candidate to the $y = x$ reference line in
$(R^{2}_{KNN},\, R^{2}_{LR})$ performance space,
i.e.\ $p_i \propto \exp(-\alpha \cdot d_i)$ with $\alpha = 10$ and
$d_i = | R^{2}_{KNN,i} - R^{2}_{LR,i} | / \sqrt{2}$.
For each candidate instance, the selection frequency (i.e. the fraction of iterations in which it
was selected) was recorded.

\begin{figure}[h]
    \centering
    \includegraphics[width=0.48\textwidth]{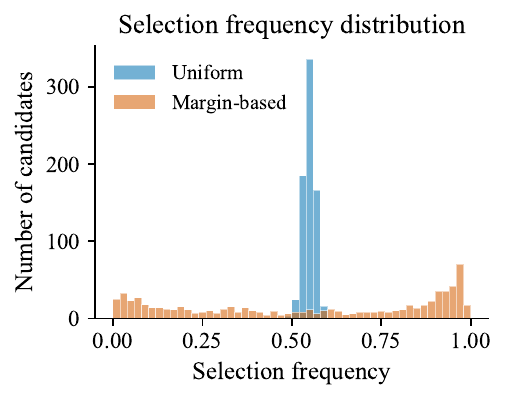}
    \caption{Distribution of per-instance selection frequencies across \num{1000}
    Monte Carlo iterations for uniform (blue) and margin-based (orange) sampling.
    Uniform sampling produces a tightly concentrated distribution around the
    expected frequency of \num{0.548}, whereas margin-based sampling spans almost
    the entire $[0, 1]$ range.}
    \label{fig:selection_frequency}
\end{figure}

\paragraph{Selection frequency distributions.}
Figure~\ref{fig:selection_frequency} presents overlapping histograms of the per-instance selection
frequencies under both strategies.
Under uniform sampling, the distribution is sharply concentrated, i.e., the mean selection frequency is
\num{0.548} with a standard deviation of only \num{0.016}, and all \num{730} candidates fall
within the narrow range $[\num{0.496},\, \num{0.590}]$.
This is consistent with the expected value of $n_{\mathrm{syn}} / n_{\mathrm{pool}} = 400/730 \approx 0.548$
and confirms that uniform sampling treats all candidates as interchangeable.
Under margin-based sampling, the mean frequency is identical (\num{0.548}, as required by the
constraint that exactly \num{400} instances are drawn per iteration), but the standard deviation
increases by more than an order of magnitude to \num{0.353}, with individual frequencies spanning
$[\num{0.005},\, \num{0.988}]$.
The margin-based histogram exhibits a different structure, with a concentration of high-frequency
candidates (those near the $y = x$ line, selected in nearly every iteration) and 
low-frequency candidates (those far from the boundary, selected rarely).
This confirms that the two strategies induce qualitatively distinct selection regimes despite
drawing the same number of instances per iteration.

% ── Figure:selection_frequency_histogram.pdf should be copied to images/ ──

\begin{figure}[h]
    \centering
    \includegraphics[width=\textwidth]{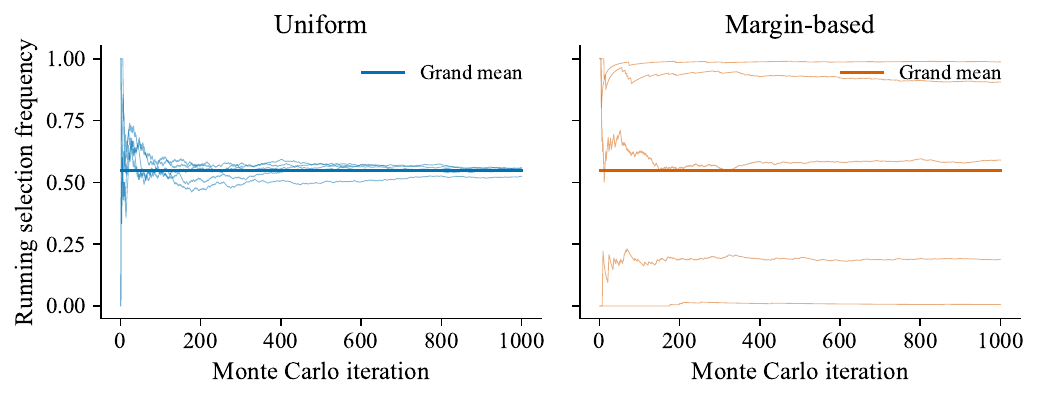}
    \caption{Convergence of running selection frequencies over \num{1000} Monte Carlo
    iterations for uniform (left) and margin-based (right) sampling. The bold line shows
    the grand mean across all \num{730} candidates; thin lines show five representative
    instances at distance percentiles (minimum, 25th, median, 75th, maximum). All
    traces stabilise well before \num{1000} iterations.}
    \label{fig:convergence}
\end{figure}

\paragraph{Convergence verification.}
Figure~\ref{fig:convergence} shows the running selection frequency as a function of the number of
Monte Carlo iterations for both strategies.
Under both uniform and margin-based sampling, individual instance traces and the grand mean converge
to stable values within approximately \num{200}--\num{300} iterations, with negligible fluctuation
thereafter.
This confirms that $N = \num{1000}$ iterations is sufficient for reliable frequency estimation, and
that the statistics reported in this analysis are not artefacts of insufficient sampling.

% ── Figure:convergence_check.pdf should be copied to images/ ──

\begin{figure}[t]
    \centering
    \includegraphics[width=0.48\textwidth]{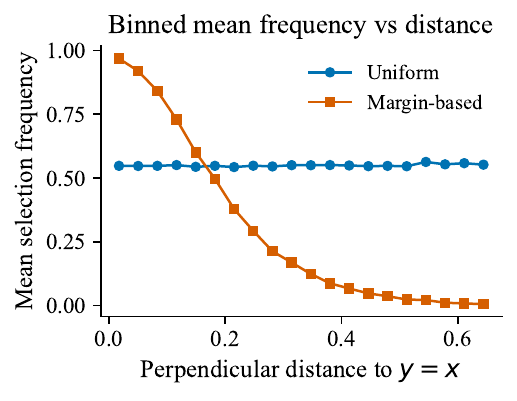}
    \caption{Binned mean selection frequency as a function of perpendicular distance to the
    $y = x$ line (\num{20} equal-width bins). The margin-based curve decreases
    monotonically from near \num{1.0} at zero distance to near \num{0.0} at the
    maximum distance, while the uniform curve remains flat.}
    \label{fig:freq_vs_distance_binned}
\end{figure}

\paragraph{Relationship between distance and selection frequency.}
The central question is whether the margin-based strategy preferentially selects instances that lie close to the $y = x$ decision boundary in performance space. Figure~\ref{fig:freq_vs_distance_binned} provides direct evidence that it does. Under uniform sampling, there is no significant linear relationship between perpendicular distance and selection frequency (Pearson $r = \num{0.063}$, $p = \num{0.088}$), confirming that distance plays no role in instance selection.
Under margin-based sampling, the correlation is strong and negative (Pearson $r = \num{-0.957}$, $p < 10^{-10}$), with instances closer to the $y = x$ line are selected with substantially higher frequency. The binned mean plot  Figure~\ref{fig:freq_vs_distance_binned}) reveals a smooth, monotonically decreasing curve for the margin-based strategy, transitioning from a mean selection frequency near \num{1.0} for candidates with distances below \num{0.05} to near \num{0.0} for candidates with distances above \num{0.5}. The uniform strategy, by contrast, produces a flat line at approximately \num{0.548} across all distance bins.

This behaviour is a direct consequence of the softmax weighting scheme, with $\alpha = 10$, the probability ratio between a candidate at the minimum observed distance ($d = \num{0.0006}$) and one at the maximum ($d = \num{0.661}$) spans approximately three orders of magnitude. From a meta-learning perspective, instances near the $y = x$ boundary are those for which the two base learners (kNN and linear regression) achieve similar predictive performance, representing ambiguous regions of the performance space where the choice of algorithm is least clear-cut. The margin-based strategy thus concentrates augmentation on precisely these difficult, boundary-proximate cases, whereas uniform augmentation distributes synthetic instances without regard for their informativeness in performance space.

\paragraph{Canonical set composition and overlap.}
To assess whether the distributional differences documented above translate into materially different instance selections, the top-\num{400} most frequently selected candidates under each strategy were identified as canonical sets. The two sets share \num{212} instances (\num{53.0}\% of each set; Jaccard similarity \num{0.361}), with \num{188} instances unique to each strategy.
This partial overlap is expected. Instances that are nearest to the $y = x$ line will be favoured by the margin-based strategy but may also appear in the uniform canonical set simply because they constitute a substantial fraction of the pool.

% ── Figure:canonical_sets_distance_distribution.pdf should be copied to images/ ──
\begin{figure}[t]
    \centering
    \includegraphics[width=0.48\textwidth]{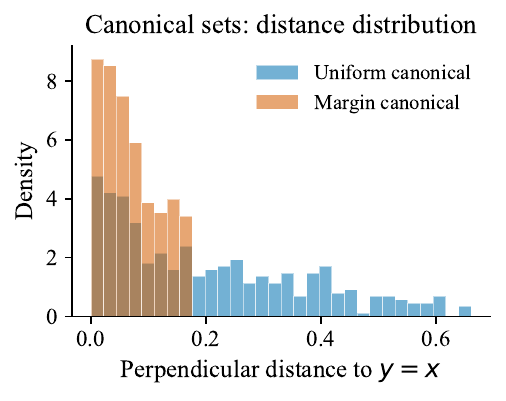}
    \caption{Normalised distribution of perpendicular distance to the $y = x$ line for the
    canonical instance sets under uniform (blue) and margin-based (orange) sampling.
    The margin-based canonical set is concentrated near $d = 0$, while the uniform set
    mirrors the pool-wide distance distribution.}
    \label{fig:canonical_distance}
\end{figure}

The critical distinction, however, lies in the distance composition of the two sets
(Figure~\ref{fig:canonical_distance}).
The margin-based canonical set has a mean perpendicular distance of \num{0.071}
(std \num{0.050}), with all \num{400} instances falling below $d = \num{0.2}$.
The uniform canonical set, by contrast, has a mean distance of \num{0.204} (std \num{0.168}),
closely mirroring the full pool distribution (mean \num{0.194}, std \num{0.161}), with only
\num{227} of \num{400} instances below $d = \num{0.2}$.
The margin-based canonical set is thus composed almost entirely of boundary-proximate instances,
whereas the uniform canonical set is an approximately representative subsample of the full
synthetic pool.
This structural difference confirms that downstream comparisons between the two canonical
sets, such as meta-feature distribution analyses, are well-motivated, as they contrast
genuinely distinct subpopulations of the synthetic candidate pool.

\paragraph{Summary.}
The Monte Carlo analysis establishes that uniform and margin-based augmentation produce
systematically and reproducibly different instance selections.
Margin-based sampling concentrates selection on instances near the $y = x$ decision
boundary in $(R^{2}_{KNN},\, R^{2}_{LR})$ performance space, with a
near-perfect inverse correlation between distance and selection frequency
($r = \num{-0.957}$), while uniform sampling is distance-agnostic.
The resulting canonical sets overlap only partially (Jaccard \num{0.361}), and differ
markedly in their distance-to-boundary composition.
These results confirm that the two strategies populate the performance space in
fundamentally different ways, providing a sound basis for the meta-feature comparison
that follows. In practical terms, this analysis validates that margin-based sampling does concentrate selection on boundary-proximate instances and that uniform sampling does provide broad coverage, confirming that the two augmentation schemes faithfully implement their intended design rationale.

\subsection{Characterisation of the Canonical Synthetic Instance Sets}
\label{sec:canonical_characterisation}

Having established that the uniform and margin-based augmentation strategies produce
systematically different instance selections (Section~\ref{sec:monte_carlo_analysis}),
we now characterise the two canonical sets of $n = \num{400}$ instances each in terms of
their meta-feature distributions and their coverage of the performance space.
Each canonical set was constructed by retaining the \num{400} most frequently selected
candidates across \num{1000} Monte Carlo iterations under the corresponding strategy.
These sets serve as the fixed representative augmentation pools for all subsequent
downstream evaluation.

\begin{figure}[!htbp]
    \centering
    \includegraphics[width=\textwidth]{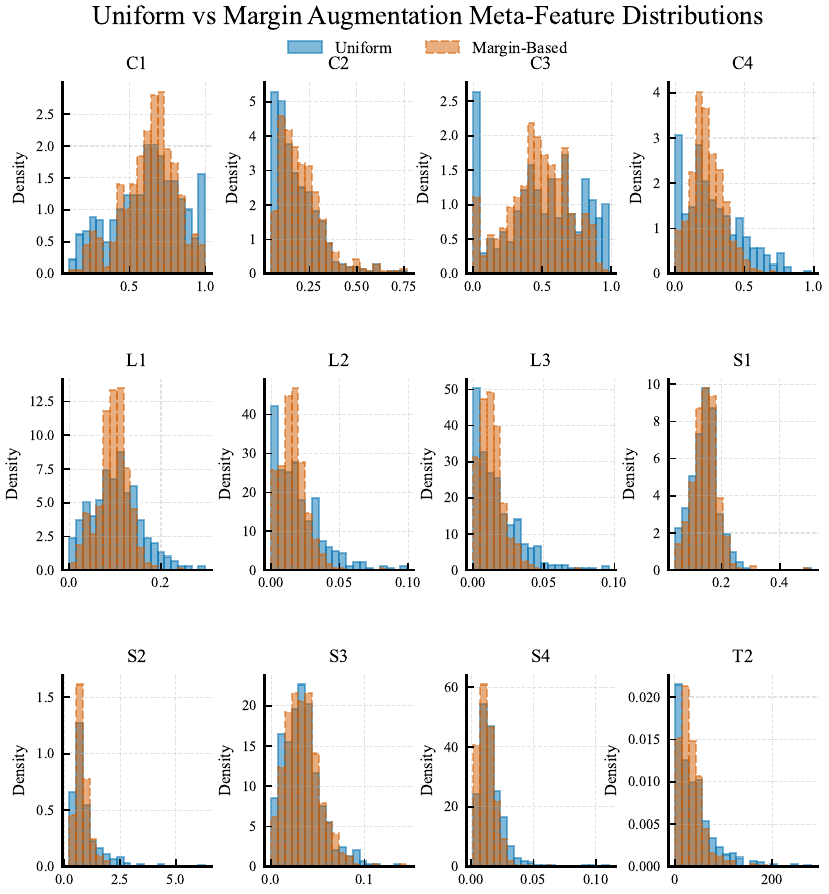}
    \caption{Density-normalised histograms of the 12 meta-features for the uniform
    (blue) and margin-based (orange) canonical instance sets (\num{400} instances each).
    Features are arranged by category: correlation and efficiency ($C1$--$C4$),
    linearity ($L1$--$L3$), smoothness ($S1$--$S4$), and dimensionality ($T2$).
    The margin-based set exhibits more concentrated distributions across the
    linearity and complexity measures, while smoothness features $S1$ and $S3$
    are effectively invariant to the augmentation strategy.}
    \label{fig:meta_feature_histograms_augmentation}
\end{figure}

\paragraph{Meta-feature distributions under uniform and margin-based augmentation.}
Figure~\ref{fig:meta_feature_histograms_augmentation} presents density-normalised
histograms of the 12 meta-features for the uniform and margin-based canonical sets.
To quantify distributional differences, we report the Kolmogorov--Smirnov (KS) test
statistic and $p$-value, Cohen's $d$ (uniform minus margin), and a histogram overlap
coefficient for each feature.
The 12 features fall into three broad sensitivity classes with respect to the augmentation strategy: features that are essentially invariant, features exhibiting
moderate shifts in location or scale, and features showing pronounced distributional
differences.

% ── Figure:meta_feature_histograms_augmentation.pdf ──

\paragraph{Invariant features.}
Two smoothness measures, $S1$ and $S3$, are effectively indistinguishable between the
two canonical sets.
For $S1$ (minimum-spanning-tree output variation), the overlap coefficient is
\num{0.928} with a negligible effect size (Cohen's $d = \num{-0.049}$, KS $p = \num{0.416}$);
means differ by only \num{0.002} (\num{0.142} vs \num{0.144}).
For $S3$ (leave-one-out 1-nearest-neighbour error), the overlap is \num{0.910} with
$d = \num{-0.007}$ and KS $p = \num{0.468}$.
These results indicate that local smoothness structure, as captured by $S1$ and $S3$,
is determined primarily by the synthetic generation process and is insensitive to
whether instances are drawn from near or far from the decision boundary in performance
space.

\paragraph{Linearity measures.}
The three linearity-related features, $L1$ (mean absolute residual), $L2$ (mean squared residual), and $L3$ (non-linearity on interpolated samples), all show highly significant
distributional differences (KS $p < 10^{-5}$ for all three).
The margin-based canonical set produces markedly more concentrated distributions:
the standard deviation of $L1$ under margin is \num{0.033} compared with \num{0.054} under
uniform; for $L2$, \num{0.010} versus \num{0.017}; and for $L3$, \num{0.009} versus
\num{0.017}.
Cohen's $d$ is positive and increasing across the trio ($d = \num{+0.160}$ for $L1$,
$\num{+0.287}$ for $L2$, $\num{+0.325}$ for $L3$), indicating that the uniform set
tends toward higher residual values on average.
The histogram shapes confirm this pattern. The margin $L2$ and $L3$ densities are
concentrated in a narrow central band, whereas the uniform densities exhibit longer
right tails extending to approximately \num{0.10} for both measures.
This concentration under margin-based selection is consistent with the strategy's
preference for boundary-proximate instances, which by construction are those on
which a linear and a nonlinear learner achieve similar performance. This implies moderate, relatively uniform residual magnitudes.

\paragraph{Complexity measures.}
Among the correlation and feature-efficiency measures $C1$--$C4$, the augmentation
strategy produces a mixed pattern of sensitivities.
$C1$ (maximum feature--target correlation) is only weakly affected, with an overlap of
\num{0.818}, $d = \num{-0.090}$, and a borderline KS $p$-value of \num{0.037}.
$C2$ (mean feature--target correlation) shifts upward under margin-based selection
(mean \num{0.212} vs \num{0.186}, $d = \num{-0.219}$, KS $p = \num{3.25e-4}$), indicating
that boundary-proximate instances tend to exhibit slightly stronger average
predictor--target associations.
$C3$ (individual feature efficiency) shows a significant KS test ($p = \num{1.76e-3}$)
but only a small effect size ($d = \num{+0.089}$); the primary difference is that
the uniform set has fatter upper tails, with density persisting above \num{0.90}
where the margin set is nearly absent (standard deviation \num{0.291} vs \num{0.219}).
$C4$ (collective feature efficiency) exhibits one of the clearest contrasts:
the uniform mean is \num{0.272} versus \num{0.233} under margin ($d = \num{+0.233}$,
KS $p = \num{1.30e-5}$), and notably the maximum value drops from \num{0.981} under
uniform to \num{0.700} under margin, indicating that the margin-based strategy
excludes the most difficult instances in terms of collective feature inefficiency.

\paragraph{Remaining sensitive features.}
The input-space smoothness measure $S2$ (input variation among neighbouring target values)
shows the largest absolute mean difference of any feature, \num{1.004} under uniform versus
\num{0.809} under margin ($d = \num{+0.306}$, KS $p = \num{3.83e-3}$).
The uniform distribution has a substantially heavier right tail, extending to a maximum of
\num{6.317} compared with \num{3.527} under margin, with the standard deviation more than
twice as large (\num{0.822} vs \num{0.365}).
$S4$ (1-nearest-neighbour non-linearity on interpolated samples) has the highest Cohen's $d$
among all 12 features ($d = \num{+0.367}$, KS $p = \num{2.29e-3}$), with the margin
distribution again being tighter (standard deviation \num{0.007} vs \num{0.012}).
Finally, the sample-to-dimension ratio $T2$ is higher and more dispersed under uniform
sampling (mean \num{41.4}, std \num{41.7}) than under margin sampling (mean \num{33.6},
std \num{26.5}; $d = \num{+0.225}$, KS $p = \num{2.29e-3}$), with the uniform maximum
reaching \num{280} compared with \num{190} under margin.

\begin{figure}[!htbp]
    \centering
    \begin{subfigure}[t]{0.48\linewidth}
        \centering
        \includegraphics[width=\linewidth]{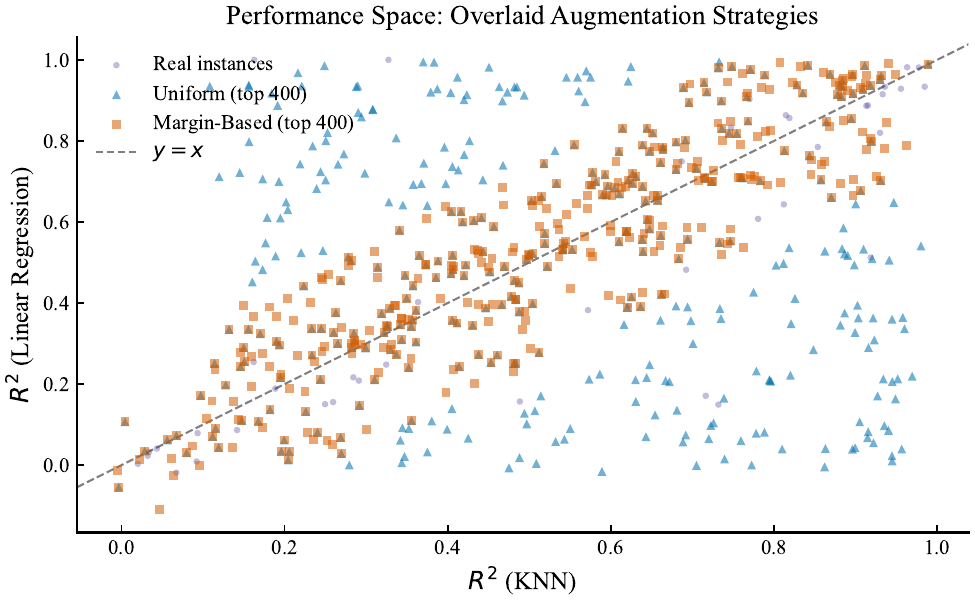}
        \caption{Scatter plot of the two canonical sets. The dashed line indicates
        the $y = x$ boundary.}
        \label{fig:perf_space_overlay}
    \end{subfigure}
    \hfill
    \begin{subfigure}[t]{0.48\linewidth}
        \centering
        \includegraphics[width=\linewidth]{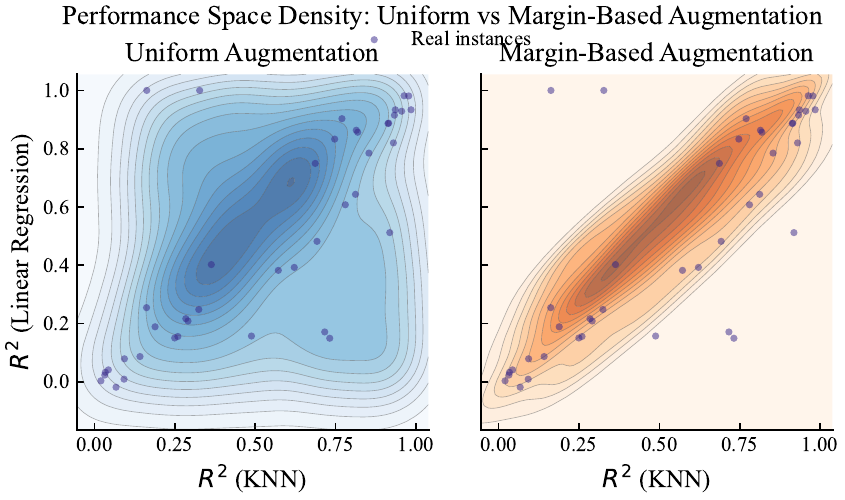}
        \caption{Kernel density estimates. The margin-based density forms a narrow
        ridge along the diagonal.}
        \label{fig:perf_space_density}
    \end{subfigure}
    \caption{Uniform (blue) and margin-based (orange) canonical instance sets in
    $(R^{2}_{KNN},\, R^{2}_{LR})$ performance space.
    (a)~The margin-based set clusters tightly along the $y = x$ diagonal, while
    the uniform set spans the full space.
    (b)~The margin-based density is concentrated in a narrow ridge, whereas the
    uniform density extends into off-diagonal regions.}
    \label{fig:perf_space}
\end{figure}

\paragraph{Performance space coverage.}
Figure~\ref{fig:perf_space_overlay} presents the two canonical sets projected into
$(R^{2}_{KNN},\, R^{2}_{LR})$ performance space, and
Figure~\ref{fig:perf_space_density} shows the corresponding density estimates.
The margin-based canonical set forms a tight band along the $y = x$ diagonal,
consistent with its construction, i.e., instances are selected in proportion to their
proximity to the line of equal algorithm performance.
The uniform canonical set, by contrast, populates a broader swath of the performance
space, including regions well above and well below the diagonal where one algorithm
substantially outperforms the other.
From an algorithm-selection perspective, the margin-based set is enriched in
\emph{ambiguous} instances, i.e., datasets for which the meta-learner must discriminate between algorithms of near-equal performance, while the uniform set includes a larger proportion
of \emph{easy} instances where the relative ordering is clear and the selection problem
is less consequential.

% ── Figures:perf_space_overlay.pdf and perf_space_density.pdf ──

\paragraph{Structural limits of LLM data synthesis.}
Three observations distil the LLM's behaviour as a generator.
First, the 484 accepted generators draw from a small family of
mechanisms whose two dials are target curvature (suppressing linear
regression) and irrelevant-feature count (suppressing KNN).
Second, the numerical vocabulary is narrow, with \texttt{np.sin}
(\num{50}\%), \texttt{np.exp} (\num{22}\%), and \texttt{np.cos}
(\num{12}\%) dominating.
Operators supporting qualitatively richer mechanisms
(\texttt{np.where}, \texttt{np.tanh}, \texttt{np.linalg.*}) appear in
under \num{2}\%.
Third, no generator invokes a domain-specific narrative (medical,
financial, physical, or otherwise) under the current prompt.
This pattern is a finding, not a failure.
More precisely, the generators occupy a bounded subspace of generation
strategies. The human-designed performance space, by contrast, is what makes their
contribution useful, since it defines where the LLM is asked to go.
This is the sense in which the structural limits of LLMs in data
synthesis are load-bearing in the pipeline rather than incidental.
These observations are elaborated in the Appendix, where the mechanism
taxonomy (Table~\ref{tab:synth_taxonomy}) and the per-generator
code-level statistics ground each claim quantitatively.

\paragraph{Summary.}
The meta-feature and performance-space analyses reveal a coherent and interpretable pattern. The margin-based augmentation strategy, by concentrating on boundary-proximate instances, produces a canonical set that is more homogeneous in meta-feature space. As such, linearity measures $L1$--$L3$ are compressed into a narrow band, the complexity measure $C4$ is capped below \num{0.700}, and the tail behaviour of $S2$, $S4$, and $T2$ is attenuated. This meta-feature compression maps directly onto the tight diagonal band observed in performance space, where both algorithms achieve similar $R^{2}$ scores. The uniform strategy, by contrast, preserves the full meta-feature diversity of the
synthetic pool, i.e., including instances with high linearity residuals, extreme feature inefficiency, and large input-space variation, and correspondingly populates a broader region of the performance space that includes clearly separable as well as ambiguous instances. The two features that are invariant to the augmentation strategy ($S1$ and $S3$) are both local smoothness measures computed via nearest-neighbour operations; their stability suggests that these properties are intrinsic to the generation mechanism rather than modulated by performance-space positioning. Whether the margin strategy's concentration on informative boundary cases or the uniform strategy's broader coverage leads to better downstream meta-learner performance is an
empirical question addressed in the evaluation that follows. The relationship between performance-space coverage and the resulting meta-feature distributions is itself not yet fully explained. Essentially, the reason why uniform sampling in performance space yields heavier-tailed meta-feature distributions while margin-based sampling yields more concentrated distributions remains an open question that warrants further investigation in future work.

\begin{table}[ht]
\centering
\caption{%
    Ablation study on data augmentation strategies across tasks.
    \emph{Uniform} sampling uniformly at random from the augmentation pool;
    \emph{Margin-based} selects augmentations near the decision boundary.
    Relative gains in the last two columns are computed as
    $\Delta = (v_{\text{new}} - v_{\text{base}})/|v_{\text{base}}|$,
    with signs reflecting the optimisation direction.
}
\label{tab:ablation}
\setlength{\tabcolsep}{7pt}
\renewcommand{\arraystretch}{1.35}
\begin{tabular}{
    l
    l
    l
    S[table-format=1.4]
    S[table-format=1.4]
    S[table-format=1.4]
    r
    r
}
\toprule
\rowcolor{headerbg}
\multicolumn{1}{l}{\color{headerfg}\textbf{Task}} &
\multicolumn{1}{l}{\color{headerfg}\textbf{Metric}} &
\multicolumn{1}{l}{\color{headerfg}\textbf{Direction}} &
\multicolumn{1}{c}{\color{headerfg}\textbf{No Aug.}} &
\multicolumn{1}{c}{\color{headerfg}\textbf{Margin}} &
\multicolumn{1}{c}{\color{headerfg}\textbf{Uniform}} &
\multicolumn{1}{c}{\color{headerfg}\textbf{Unif.\ vs\ Margin}} &
\multicolumn{1}{c}{\color{headerfg}\textbf{Unif.\ vs\ No Aug.}} \\
\midrule

\rowcolor{subheadbg}
\multicolumn{7}{l}{\textbf{Multi-label Classification}} \\[1pt]

& Hamming Loss     & $\downarrow$ & 0.4050 & 0.3531 & \best{0.3342} & \poscolor{+5.34\%}   & \poscolor{+17.47\%}  \\
& Subset Acc.      & $\uparrow$   & 0.1752 & 0.2598 & \best{0.3512} & \poscolor{+35.20\%}  & \poscolor{+100.41\%} \\

\midrule

\rowcolor{subheadbg}
\multicolumn{7}{l}{\textbf{Regression}} \\[1pt]

& Pooled OOF $R^2$ & $\uparrow$   & 0.7646 & 0.7940 & \best{0.8112} & \posweak{+2.17\%}    & \posweak{+6.09\%}    \\

\bottomrule
\end{tabular}
\end{table}

\subsection{Ablation Analysis and Learning Curves}
\label{sec:ablation_learning_curves}

\paragraph{Ablation study.}
Table~\ref{tab:ablation} summarises performance at a fixed augmentation budget of $n_{\mathrm{syn}} = \num{400}$ across both downstream meta-learning formulations. Three conditions are compared: no augmentation (the real-world meta-dataset alone),
margin-based augmentation, and uniform augmentation. The ablation isolates the effect of the augmentation strategy while holding the
augmentation volume and evaluation protocol constant, and therefore provides a direct test of H2, comparing margin-based concentration against uniform coverage. Across all three metrics, uniform augmentation achieves the best performance, and both augmentation strategies substantially outperform the no-augmentation baseline.

\paragraph{Multi-label classification.}
For the multi-label formulation, uniform augmentation reduces Hamming loss from \num{0.4050} (no augmentation) to \num{0.3342}, a relative improvement of \num{17.47}\%. Margin-based augmentation also improves over the baseline (\num{0.3531}), but uniform sampling outperforms it by a further \num{5.34}\%. The contrast is substantially sharper on subset accuracy, where uniform augmentation achieves \num{0.3512}, a \num{100.41}\% relative gain over the no-augmentation baseline of \num{0.1752} and a \num{35.20}\% gain over margin-based augmentation (\num{0.2598}).
The magnitude of this improvement is notable, with uniform augmentation doubling the exact-match recovery rate of the meta-learner's predicted label vectors.
This finding is reinforced by the H1 endpoint paired t-tests at $n_{\mathrm{syn}} = \num{730}$, where augmentation differs from the no-augmentation baseline at $p < 10^{-14}$ for both Hamming loss (mean difference $-\num{0.078}$) and subset accuracy (mean difference $+\num{0.151}$).
The large gap between Hamming loss improvement (\num{17.47}\%) and subset accuracy improvement (\num{100.41}\%) indicates that uniform augmentation is particularly effective at stabilising the collective relationship between algorithms. This suggests that broad coverage helps the meta-learner resolve the entire performance topography required for exact-match recommendations, whereas narrow sampling may leave the joint label structure under-defined.

It is important to note that the relatively low absolute subset accuracy ($\approx$\num{0.35}) reflects the inherent stringency of the metric rather than poor model quality. Because the label vector has $K=5$ binary components, a prediction is counted as correct only if all five labels match exactly; a prediction that correctly identifies four out of five applicable algorithms still counts as a complete miss. The combinatorial space of $2^5 = 32$ possible label vectors makes exact-match recovery substantially harder than per-label prediction, which is why Hamming loss presents a more favourable picture of the same underlying model.

\paragraph{Regression.}
In the regression formulation, both augmentation strategies improve pooled OOF $R^{2}$ over the no-augmentation baseline of \num{0.7646}. Uniform augmentation achieves \num{0.8112} (\num{+6.09}\% relative), while margin-based augmentation reaches \num{0.7940} (\num{+3.84}\% relative).
This finding is reinforced by the H1 endpoint paired t-test at $n_{\mathrm{syn}} = \num{730}$, where augmentation differs from the no-augmentation baseline at $p < 10^{-15}$ (mean difference $+\num{0.057}$ in $R^{2}$).
The gap between the two strategies is smaller here (\num{+2.17}\%) than for either multi-label metric, indicating that the SVR-based regression meta-learner is less sensitive to augmentation strategy choice than the Naive Bayes classifier. This is consistent with the observation in Section~\ref{sec:canonical_characterisation} that the two canonical sets differ most markedly in meta-feature dispersion
(linearity measures, complexity $C4$, and tail behaviour of $S2$, $S4$, $T2$); the regression meta-learner, which operates on continuous performance targets, appears more robust to these distributional differences than the multi-label classifier, which must recover discrete label boundaries.

The pooled OOF $R^2$ of \num{0.8112} under uniform augmentation further contextualises the low subset accuracy. Essentially, the underlying continuous performance predictions are reasonably accurate, and the multi-label classifier's low exact-match rate is a consequence of discretising these predictions into binary labels and then requiring all five to be simultaneously correct, rather than an indication of poor meta-learner quality.

\paragraph{Strategy comparison.}
The consistent advantage of uniform over margin-based augmentation across all three metrics suggests that the meta-learner prioritises a \textit{comprehensive mapping of the performance landscape} over localised refinement.
The H2 paired t-tests run at every learning-curve point reflect this advantage. Across the testable train sizes per metric, uniform augmentation exceeds margin-based augmentation at $p < 0.05$ for 34 of 37 on $R^{2}$, 28 of 36 on Hamming loss, and 30 of 36 on subset accuracy. The few reversals favouring margin-based augmentation are confined to a narrow window near $n_{\mathrm{syn}} = \num{560}$--$\num{580}$ on Hamming loss and a single point at $n_{\mathrm{syn}} = \num{580}$ on subset accuracy.
While the margin-based strategy was designed to concentrate on the most ambiguous instances, the results indicate that generalisation is primarily driven by \textit{establishing a global structural foundation} that prevents predictive gaps in the algorithm space. Three factors may contribute to this outcome. First, the temperature parameter $\alpha = 10$ used for margin-based sampling produces a relatively broad sampling distribution
(Section~\ref{sec:monte_carlo_analysis}): at this setting, the canonical margin set still contains instances with distances up to $d = \num{0.2}$, rather than being tightly concentrated on the boundary. Second, and more fundamentally, the meta-learners evaluated here may benefit more from the distributional diversity provided by uniform coverage than from the concentrated boundary signal provided by margin-based selection.
Third, margin-based sampling collapses the joint variation of $R^{2}_{KNN}$ and $R^{2}_{LR}$ within the augmented meta-dataset. By drawing only synthetic rows whose two landmarker scores agree, it forces these coordinates into near-collinearity across the appended training rows. The performance space is a partial projection of the meta-feature representation when landmarking information is included (Section~\ref{sec:hypotheses}), so this collinearity reduces the independent information that the two coordinates contribute to the meta-learner. Uniform sampling, by contrast, preserves the full range of $(R^{2}_{KNN},\, R^{2}_{LR})$ pairs and keeps the two coordinates approximately uncorrelated.
The ablation evidence supports the conclusion that global anchoring provides a more robust scaffold for the meta-learner than targeted boundary sampling at moderate budgets. This confirms that a distributed cover of the performance space is a prerequisite for generalisation, ensuring that the meta-learner is not biased by local clusters before the full range of algorithmic behaviours has been represented.

\begin{figure}[!htbp]
    \centering
    \begin{subfigure}[t]{0.48\linewidth}
        \centering
        \includegraphics[width=\linewidth]{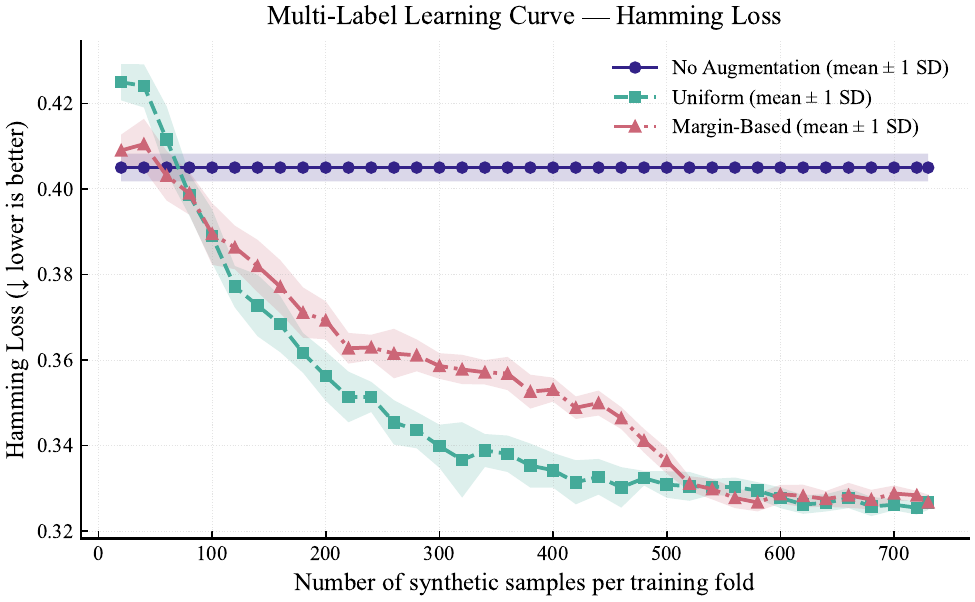}
        \caption{Hamming loss (lower is better). Both augmentation strategies
        initially degrade performance at $n_{\mathrm{syn}} = \num{20}$ before
        crossing below the baseline near $n_{\mathrm{syn}} \approx \num{60}$--$\num{80}$.}
        \label{fig:lc_hamming}
    \end{subfigure}
    \hfill
    \begin{subfigure}[t]{0.48\linewidth}
        \centering
        \includegraphics[width=\linewidth]{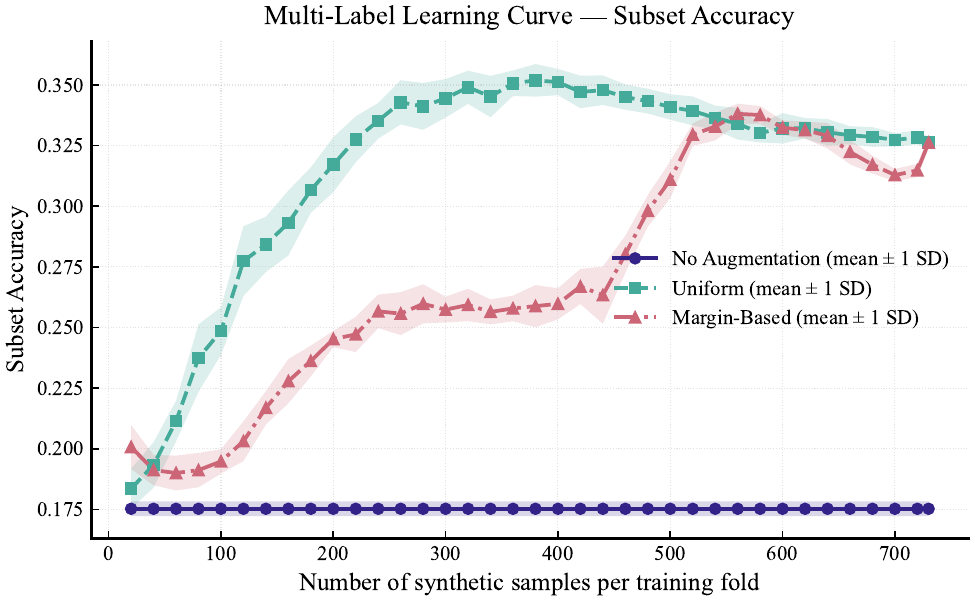}
        \caption{Subset accuracy (higher is better). Uniform augmentation peaks
        near $n_{\mathrm{syn}} \approx \num{380}$ and subsequently declines,
        exhibiting a non-monotonic trajectory.}
        \label{fig:lc_subset}
    \end{subfigure}
    \caption{Multi-label classification learning curves as a function of the
    number of synthetic samples per training fold.
    Each curve reports the mean across \num{10} seeds; shaded bands show
    $\pm 1$ standard deviation.
    The no-augmentation baseline (flat line) is included for reference.}
    \label{fig:lc_multilabel}
\end{figure}

\begin{figure}[!htbp]
    \centering
    \includegraphics[width=0.48\textwidth]{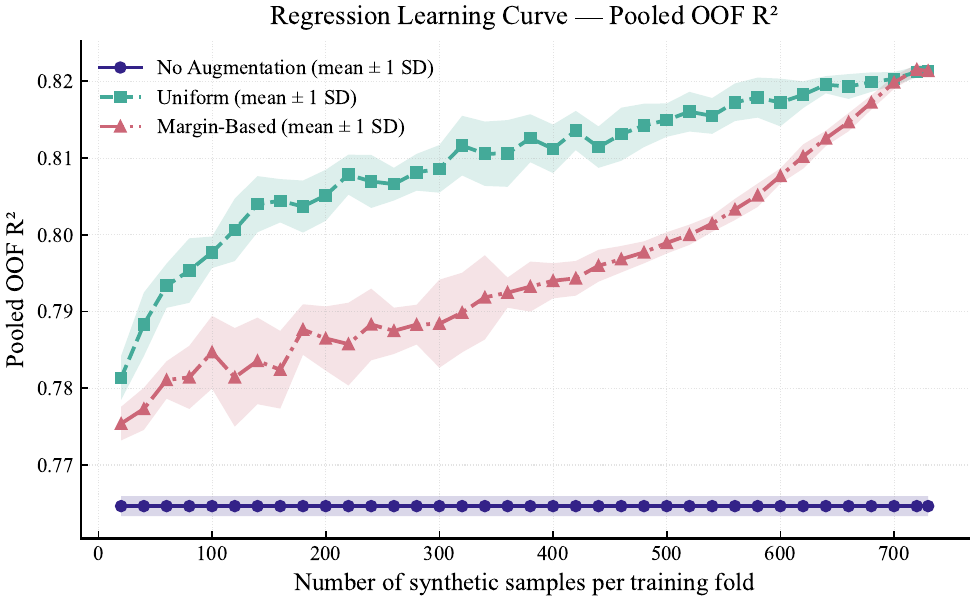}
    \caption{Regression learning curve for pooled OOF $R^{2}$ (higher is better)
    as a function of the number of synthetic samples per training fold.
    Unlike the multi-label setting, both augmentation strategies improve over
    the baseline immediately at $n_{\mathrm{syn}} = \num{20}$ and rise
    monotonically throughout the observed range.
    Mean $\pm 1$ standard deviation across \num{10} seeds.}
    \label{fig:lc_regression}
\end{figure}

\paragraph{Learning curve analysis.}
To examine how meta-learner performance evolves as the augmentation volume increases, learning curves were computed by sweeping the number of synthetic samples per training fold from $n_{\mathrm{syn}} = \num{20}$ to \num{730} (the full synthetic pool) in steps of \num{20}. At each budget, the same repeated cross-validation protocol ($K = 10$, \num{10} repeats, \num{10} seeds) was applied under the no-augmentation, uniform, and margin-based conditions. Figures~\ref{fig:lc_multilabel} and~\ref{fig:lc_regression} present the resulting curves, with shaded bands indicating $\pm 1$ standard deviation across seeds.

\begin{figure}[!htbp]
    \centering
    \includegraphics[width=0.6\textwidth]{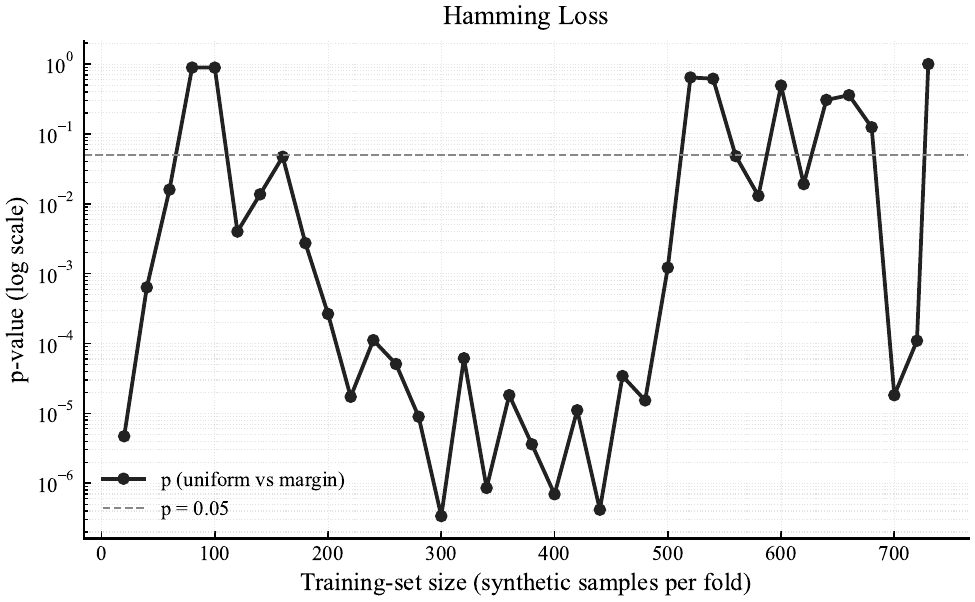}
    \caption{H2 paired t-test $p$-values for Hamming loss as a function of
    the number of synthetic samples per training fold. The dashed line at
    $\alpha = \num{0.05}$ marks the significance threshold; the sign of the
    mean difference (uniform $-$ margin) indicates the direction of the effect.}
    \label{fig:pvalh2_hamming}
\end{figure}

\paragraph{Hamming loss learning curve.}
The Hamming loss curve (Figure~\ref{fig:lc_hamming}) reveals a cold-start effect. At $n_{\mathrm{syn}} = \num{20}$, both uniform (\num{0.425}) and margin-based (\num{0.409}) augmentation produce \emph{worse} Hamming loss than the no-augmentation baseline (\num{0.405}). This initial degradation reflects the noise introduced by a small number of synthetic instances into the Naive Bayes classifier's conditional density estimates. From a practical standpoint, this cold-start effect implies that practitioners should avoid very small augmentation budgets. A minimum of approximately $n_{\mathrm{syn}} \geq \num{80}$ synthetic datasets is needed before augmentation begins to outperform the unaugmented baseline for this meta-learner.
Uniform augmentation crosses below the baseline near $n_{\mathrm{syn}} \approx \num{80}$ and then decreases steeply, reaching a plateau around $n_{\mathrm{syn}} \approx \num{460}$--$\num{500}$ at approximately \num{0.330}. Margin-based augmentation follows a similar trajectory but with a shallower
descent, plateauing later near $n_{\mathrm{syn}} \approx \num{560}$--$\num{580}$. Uniform is consistently lower (better) than margin-based across the mid-range ($n_{\mathrm{syn}} = \num{80}$--$\num{520}$). At $n_{\mathrm{syn}} = \num{730}$, both curves converge to an identical Hamming loss of \num{0.327}, as expected when the full synthetic pool is used under both strategies.
This pattern is reinforced by the H2 paired t-tests (Figure~\ref{fig:pvalh2_hamming}). Uniform augmentation differs from margin-based augmentation at $p < 0.05$ for 28 of 36 testable train sizes. Most significant differences in the mid-range ($n_{\mathrm{syn}} = \num{120}$--$\num{500}$) favour uniform augmentation; the brief reversals at $n_{\mathrm{syn}} = \num{560}$ ($p = \num{0.048}$) and $n_{\mathrm{syn}} = \num{580}$ ($p = \num{0.013}$) favour margin-based augmentation by a small mean difference of about \num{0.003}. At $n_{\mathrm{syn}} = \num{730}$ the paired samples coincide and no test is computed.

\paragraph{Subset accuracy learning curve.}

The subset accuracy curve (Figure~\ref{fig:lc_subset}) exhibits a qualitatively different and non-monotonic pattern. Uniform augmentation rises steeply from \num{0.184} at $n_{\mathrm{syn}} = \num{20}$, peaks at \num{0.352} near $n_{\mathrm{syn}} \approx \num{380}$, and then \emph{declines} to \num{0.326} by $n_{\mathrm{syn}} = \num{730}$. Margin-based augmentation rises more gradually, reaching a peak of \num{0.338} at $n_{\mathrm{syn}} \approx \num{560}$, before a similar decline. The peak-then-decline trajectory is notable because it contrasts with the monotonically improving Hamming loss over the same range of $n_{\mathrm{syn}}$. Since Hamming loss measures marginal per-label accuracy while subset accuracy measures exact-match recovery of the full label vector, the divergence suggests that at high augmentation volumes the classifier improves on individual label decisions but becomes less calibrated in their joint prediction---producing predictions that are marginally more accurate but combinatorially less precise. The decline in subset accuracy for $n_{\mathrm{syn}} > \num{400}$ suggests a \textit{saturation point} where the synthetic distribution begins to overshadow the unique characteristics of real-world datasets. This implies that once the \textit{structural grid} is sufficiently established, adding further synthetic volume may introduce noise that destabilises the fine-grained coordination of the algorithm labels.

The decline region ($n_{\mathrm{syn}} > \num{400}$) also coincides with the regime in which synthetic instances begin to outnumber real-world instances in each training fold, suggesting that over-representation of the synthetic distribution may erode the meta-learner's ability to recover the joint label structure of real-world datasets. For practitioners, this non-monotonic behaviour suggests that subset accuracy should be monitored alongside Hamming loss when tuning the augmentation budget, and that an intermediate budget (around $n_{\mathrm{syn}} \approx \num{380}$--$\num{400}$) may offer the best trade-off between per-label and joint-label prediction quality.

\begin{figure}[!htbp]
    \centering
    \includegraphics[width=0.6\textwidth]{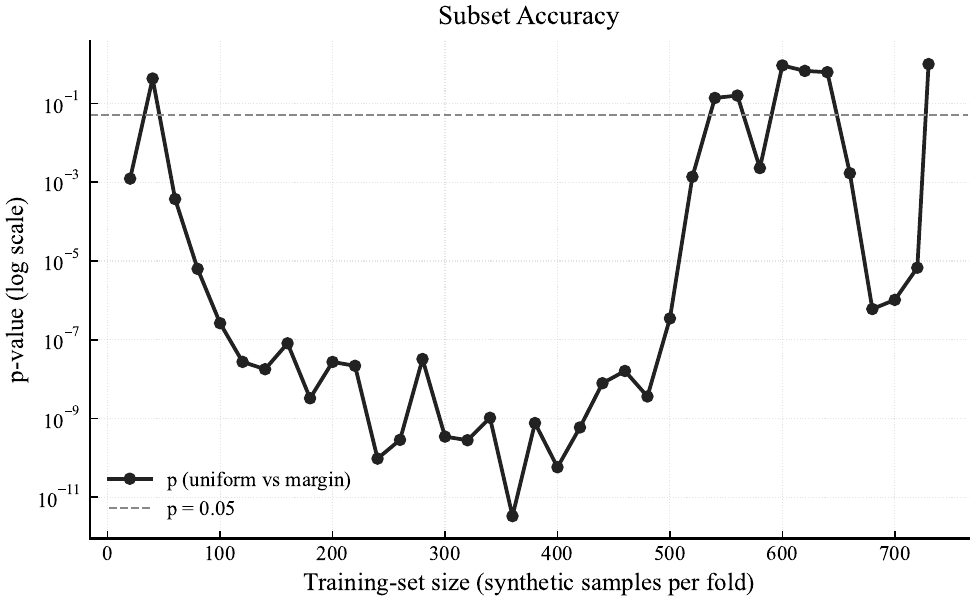}
    \caption{H2 paired t-test $p$-values for subset accuracy as a function of
    the number of synthetic samples per training fold. The dashed line at
    $\alpha = \num{0.05}$ marks the significance threshold; the sign of the
    mean difference (uniform $-$ margin) indicates the direction of the effect.}
    \label{fig:pvalh2_subset}
\end{figure}

The eventual convergence of both strategies at $n_{syn}=730$ suggests that the superiority of uniform sampling is a function of structural completeness. We hypothesise that the meta-learning process undergoes a functional shift, i.e., an initial phase where broad spatial anchoring is required to define the performance manifold, followed by a state where the manifold is sufficiently "populated" that additional global samples yield diminishing returns compared to boundary-focused data. We hypothesise that the meta-learner undergoes a phase transition: an initial "expansion phase" where global coverage (Uniform) is required to anchor the structural scaffolding of the performance space, followed by a "refinement phase" where margin-adjacent samples become more informative. The observed cross-over points in the Hamming loss and $R^2$ curves around $n_{syn} \approx 600$ provide preliminary evidence for this \textit{Hybrid-Switch Hypothesis}, suggesting that the ideal augmentation trajectory is not static but should pivot from uniform to margin-based sampling as the augmentation volume increases.
This pattern is reinforced by the H2 paired t-tests (Figure~\ref{fig:pvalh2_subset}). Uniform augmentation differs from margin-based augmentation at $p < 0.05$ for 30 of 36 testable train sizes, with the bulk of significant differences in the mid-range ($n_{\mathrm{syn}} = \num{60}$--$\num{520}$) favouring uniform augmentation. The cold-start point at $n_{\mathrm{syn}} = \num{20}$ favours margin-based augmentation (mean difference $-\num{0.017}$, $p = \num{0.0012}$), and a brief reversal at $n_{\mathrm{syn}} = \num{580}$ ($p = \num{0.002}$) again favours margin-based augmentation by about \num{0.007}. At $n_{\mathrm{syn}} = \num{730}$ the paired samples coincide and no test is computed.

\begin{figure}[!htbp]
    \centering
    \includegraphics[width=0.6\textwidth]{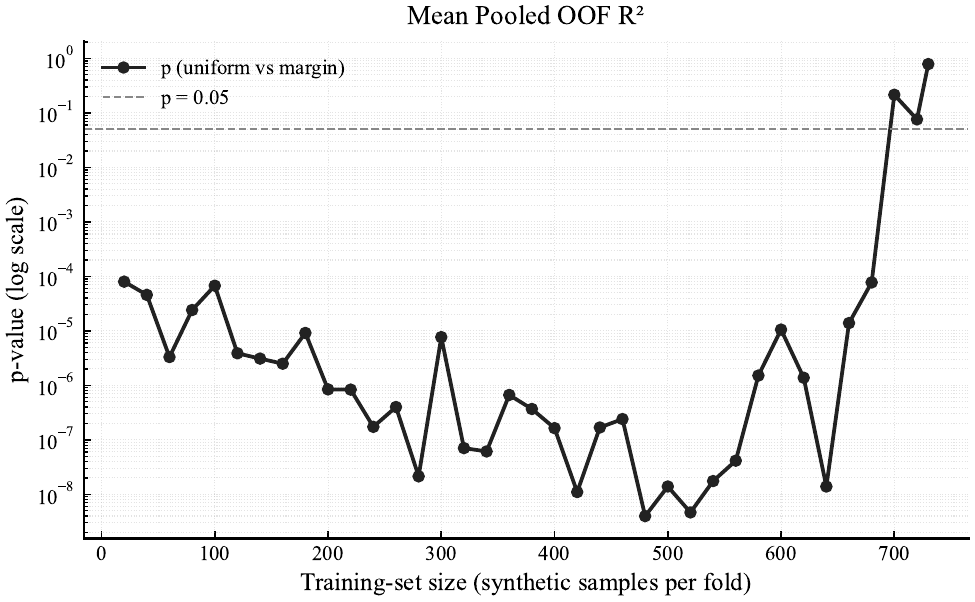}
    \caption{H2 paired t-test $p$-values for pooled OOF $R^{2}$ as a function of
    the number of synthetic samples per training fold. The dashed line at
    $\alpha = \num{0.05}$ marks the significance threshold; the sign of the
    mean difference (uniform $-$ margin) indicates the direction of the effect.}
    \label{fig:pvalh2_r2}
\end{figure}

\paragraph{Regression learning curve.}
The pooled OOF $R^{2}$ curve (Figure~\ref{fig:lc_regression}) contrasts with the multi-label setting in two respects.
First, there is no cold-start degradation. Both augmentation strategies improve over the no-augmentation baseline (\num{0.765}) immediately at $n_{\mathrm{syn}} = \num{20}$ (uniform: \num{0.781}; margin: \num{0.775}). Second, the improvement is monotonically increasing across the entire observed
range, with no peak-then-decline behaviour. Uniform augmentation rises steadily to \num{0.821} at $n_{\mathrm{syn}} = \num{730}$,
while margin-based augmentation follows a slower trajectory, narrowing the gap only after $n_{\mathrm{syn}} \approx \num{600}$ and converging to the same value at $n_{\mathrm{syn}} = \num{730}$. The absence of a cold-start effect is consistent with the SVR meta-learner's kernel-based formulation, which is inherently more robust to small perturbations in the training distribution than the density-estimation-based Naive Bayes
classifier. Error bands narrow substantially with increasing $n_{\mathrm{syn}}$ (standard deviation across seeds drops from approximately \num{0.003} at $n_{\mathrm{syn}} = \num{20}$ to \num{0.0006} at $n_{\mathrm{syn}} = \num{730}$), indicating that larger augmentation budgets stabilise meta-learner training across random seeds and cross-validation partitions.
The above pattern is again reinforced by the H2 paired t-tests (Figure~\ref{fig:pvalh2_r2}). Uniform augmentation exceeds margin-based augmentation at $p < 0.05$ for 34 of the 37 testable train sizes, with mean differences in $R^{2}$ ranging from $+\num{0.006}$ at $n_{\mathrm{syn}} = \num{20}$ to a peak of $+\num{0.022}$ around $n_{\mathrm{syn}} \approx \num{220}$--$\num{320}$ before contracting toward the convergence point.
The non-significant points are confined to $n_{\mathrm{syn}} = \num{700}$ ($p = \num{0.215}$), $n_{\mathrm{syn}} = \num{720}$ ($p = \num{0.076}$), and $n_{\mathrm{syn}} = \num{730}$ ($p = \num{0.788}$), all at the upper end where the two strategies converge.

\paragraph{Cross-curve comparison.}
Three patterns emerge from the joint analysis of the three learning curves. 
First, uniform augmentation dominates margin-based augmentation across essentially the entire $n_{\mathrm{syn}}$ range for all three metrics. Margin-based augmentation matches or overtakes uniform augmentation only in narrow windows. More specifically, for Hamming loss significant reversals appear at $n_{\mathrm{syn}} = \num{560}$ ($p = \num{0.048}$) and $n_{\mathrm{syn}} = \num{580}$ ($p = \num{0.013}$), with mean differences below \num{0.003}. For subset accuracy, the reversal at $n_{\mathrm{syn}} = \num{580}$ is also significant ($p = \num{0.002}$, mean difference about $-\num{0.007}$). For $R^{2}$, the apparent dip at $n_{\mathrm{syn}} = \num{720}$ is not statistically significant ($p = \num{0.076}$). None of these reversals alter the overall ranking.

Second, at $n_{\mathrm{syn}} = \num{730}$, both strategies necessarily draw the entire synthetic pool, so any remaining difference would indicate an implementation artefact rather than a strategy effect. The observed convergence to identical terminal scores across all three metrics confirms that the mid-range performance gaps are genuinely attributable to how the two strategies sample different subsets of the pool, and serves as a consistency check on the experimental pipeline. Third, the variance-reduction effect is universal, i.e., across all three metrics, error bands contract monotonically with $n_{\mathrm{syn}}$, demonstrating that synthetic augmentation improves not only average performance but also the reproducibility of meta-learner evaluation.

The ablation table (Table~\ref{tab:ablation}) reports results at
$n_{\mathrm{syn}} = \num{400}$, which the learning curves reveal to be situated near the optimum for uniform subset accuracy (peak at $n_{\mathrm{syn}} \approx \num{380}$) but well before the plateau for Hamming loss ($n_{\mathrm{syn}} \approx \num{460}$--$\num{500}$) and in the middle of the monotonic ascent for $R^{2}$. The metric for which the ablation table reports the largest strategy gap, subset accuracy, which is \num{+35.20}\% for uniform over margin, is also the metric whose learning curves show the widest separation between the two strategies in the mid-range, confirming that the fixed-budget comparison reflects a genuine and sustained performance difference rather than an artefact of the particular $n_{\mathrm{syn}}$ chosen, which the results of our elaborate hypothesis testing over the entire learning curve also supports.

\paragraph{Summary.}
The ablation and learning curve analyses jointly provide a detailed picture of how synthetic data augmentation affects meta-learner behaviour. The ablation study confirms that both augmentation strategies substantially improve performance over the baseline, which only includes real-world datasets. The H1 endpoint paired t-tests show $p < 10^{-14}$ for all three in-scope metrics, with mean differences of $+\num{0.057}$ for $R^{2}$, $-\num{0.078}$ for Hamming loss, and $+\num{0.151}$ for subset accuracy, confirming H1's alternate. Beyond the baseline comparison, uniform augmentation outperforms margin-based augmentation not only at the fixed budget but across the learning curve. The H2 paired t-tests are significant at $\alpha = \num{0.05}$ for 34 of 37 train sizes on $R^{2}$, 28 of 36 on Hamming loss, and 30 of 36 on subset accuracy, supporting H2's alternate.
The learning curves add three further insights: (i) the benefit of augmentation is dose-dependent but subject to diminishing returns, with most of the gain captured by $n_{\mathrm{syn}} \approx \num{400}$--$\num{500}$; (ii) the multi-label meta-learner exhibits a cold-start penalty and a non-monotonic subset accuracy trajectory that are absent in the regression meta-learner,
highlighting the sensitivity of the Naive Bayes classifier to the
synthetic-to-real-world data ratio; and (iii) the advantage of uniform over margin-based augmentation is not an artefact of the fixed budget but persists across the full range of augmentation volumes, narrowing only as $n_{\mathrm{syn}}$ approaches the total synthetic pool size. These results suggest that, for the meta-learners and task representations evaluated here, distributional diversity in the augmentation set is more beneficial than concentrated boundary signal, and that augmentation budgets should be chosen with attention to the specific metric of interest, given the divergent behaviour of Hamming loss and subset accuracy at high volumes.

A notable exception to the monotonic improvement trend is the Naive Bayes classifier, which exhibits a "\textit{cold-start}" effect, i.e., a performance dip at $n_{syn }=20$. This suggests that sparse, noisy synthetic samples may initially interfere with the learner’s density estimation. This phenomenon indicates that for meta-learning tasks, a critical mass or minimum synthetic-to-real-world ratio (approximately 1:5 in this context) may be required before the benefits of augmentation outweigh the noise introduced by the generative process. Practitioners should therefore ensure that synthetic batches are of sufficient scale to represent the variance of the target performance distribution rather than just its mean.

While both strategies eventually meet at $n_{syn}=730$ (where the sampling pools become identical), the trajectory leading to this point reveals a significant \textit{strategy crossover} around $n_{syn} \approx 600$. The initial dominance of uniform sampling suggests that broad spatial anchoring is required to define the structural scaffolding of the performance manifold in early learning stages. However, as the Hamming loss and $R^2$ curves intersect, we observe a pivot in relative utility, i.e., that the marginal benefit of global coverage diminishes, and the learner begins to derive more value from margin-adjacent refinement. This empirical shift suggests that the optimal augmentation strategy is volume-dependent, transitioning from a focus on \textit{topological covering} to boundary precision as the synthetic budget increases.

\paragraph{Coverage-granularity comparison.}
As a preliminary check on top of the uniform-augmentation benefit established above, we compare the 5x5 and 7x7 grids in isolation, with each grid used in turn as the full augmentation pool and pairs formed by seed. This is not a new hypothesis test, only a granularity probe at the two grid resolutions available in this work. Table~\ref{tab:granularity} reports the paired-difference results. All three metrics show statistically significant per-seed differences, but the directions disagree. Pooled OOF $R^{2}$ favours the 7x7 grid, while Hamming Loss and Subset Accuracy both favour the 5x5 grid. The regression metric and the two classification metrics therefore split, providing inconclusive evidence of a clear winner at this single grid pair.

\begin{table}[ht]
\centering
\caption{%
    Coverage-granularity paired comparison between the 7x7 grid
    (\num{482} synthetic datasets) and the 5x5 grid (\num{248} synthetic datasets),
    each used in turn as the full augmentation pool.
    Pairs are formed across $n = \num{10}$ seeds.
    Mean differences are 7x7 minus 5x5, with two-sided paired Student's $t$-tests.
    The rightmost column lists the grid favoured by the sign of the mean
    difference under each metric's optimisation direction.
}
\label{tab:granularity}
\setlength{\tabcolsep}{8pt}
\renewcommand{\arraystretch}{1.3}
\begin{tabular}{
    l
    c
    S[table-format=+1.4]
    l
    l
    c
}
\toprule
\rowcolor{headerbg}
\multicolumn{1}{l}{\color{headerfg}\textbf{Metric}} &
\multicolumn{1}{c}{\color{headerfg}\textbf{Direction}} &
\multicolumn{1}{c}{\color{headerfg}\textbf{Mean $\Delta$}} &
\multicolumn{1}{c}{\color{headerfg}\textbf{95\% CI}} &
\multicolumn{1}{c}{\color{headerfg}\textbf{$p$-value}} &
\multicolumn{1}{c}{\color{headerfg}\textbf{Favours}} \\
\midrule
Pooled OOF $R^{2}$ & $\uparrow$   & +0.0166 & $[+\num{0.0163},\,+\num{0.0170}]$ & $\num{7.4e-15}$ & 7x7 \\
Hamming Loss       & $\downarrow$ & +0.0065 & $[+\num{0.0038},\,+\num{0.0092}]$ & $\num{4.0e-4}$  & 5x5 \\
Subset Accuracy    & $\uparrow$   & -0.0202 & $[-\num{0.0230},\,-\num{0.0174}]$ & $\num{5.5e-8}$  & 5x5 \\
\bottomrule
\end{tabular}
\end{table}

\subsection{Structural Implications: The Geometry of Generalisation}\label{sec:witness_logic}

The consistent advantage of uniform- over margin-based sampling suggests that meta-learning is not merely about identifying decision boundaries, but is instead a process of \textit{reconstructing the functional topography} of the entire algorithm space. This section interprets these results through a structural framework that centres on the relationship between synthetic data and the geometric manifold of real-world dataset performance.

\paragraph{The Performance Manifold and Universal Addressing.}
We define a \textit{structural embedding} as a mapping where every dataset $D$ is assigned a "universal address" based on its performance characteristics. This address exists within a \textit{landmarker performance space} (LPS), which corresponds to a geometric container where the axes represent the performance of simple "landmarker" algorithms (e.g., KNN or LR). In this study, the LPS is a 2-dimensional space defined by:
$$\phi(D) = (R^2_{KNN}(D), R^2_{LR}(D)) \in [0, 1]^2$$\\ This mapping creates a \textit{performance manifold}, which is a lower-dimensional projection of the broader dataset space. Our results suggest that while the final meta-learning target is 5-dimensional, the underlying structural logic of the datasets is effectively captured within this 2-dimensional LPS.

\paragraph{Topological Covering and Resolution Matching}
The primary implication of our findings is that the meta-learner’s success depends on \textit{topological covering}, i.e., the requirement that synthetic data points are spread across the manifold to leave no significant gaps. This is formalised through the $\epsilon$\textit{-cover} condition:

\begin{center}
%\rev{A synthetic meta-dataset $\mathcal{S}$ provides an $\epsilon$-cover of the space of real-world datasets $\mathcal{X}$ if for every real-world dataset, there is a synthetic "witness" within a distance of $\epsilon$ in the LPS.}
A synthetic meta-dataset $\mathcal{S}$ provides an $\epsilon$-cover of the space of real-world datasets $\mathcal{X}$ if: $$\forall x \in \mathcal{X}, \exists s \in \mathcal{S} : \|\phi(x) - \phi(s)\|_2 < \epsilon$$
\end{center}
This condition ensures that the synthetic training data effectively brackets the performance characteristics of any potential real-world dataset.

Uniform sampling outperforms margin-based methods because it minimises \textit{manifold reconstruction bias}. While margin-based sampling focuses on "ambiguous" areas, uniform sampling ensures a global cover that prevents "blind spots" in the manifold reconstruction. This process enables \textit{resolution matching}, where the "grain" or density of the synthetic datasets matches the inherent structural resolution of the performance manifold.

%\rev{However, the dominance of this $\epsilon$-cover condition is inherently phase-dependent. we propose a \textit{Manifold Maturity Threshold} ($\tau$), defined as the density at which the global topological reconstruction provides sufficient stability to transition from "expansion" to "specialisation". In this framework, the uniform $\epsilon$-cover is a temporal prerequisite, i.e., it provides the global anchoring that prevents the meta-learner from overfitting to local clusters before the full functional topography is mapped. Once the maximum distance to a synthetic witness falls below $\tau$, the marginal utility of uniform sampling diminishes, and the focus must pivot toward boundary-sharpening.}

\paragraph{Steerable Synthesis and the Propose-Execute-Repair Loop.}
The ability to achieve this cover within a finite computational budget is driven by \textit{steerable synthesis}, i.e., the capacity of an LLM to generate datasets with specific, targeted properties. This is operationalized through a \textit{Propose-Execute-Repair Loop}:

\begin{itemize}
    \item Propose: The LLM suggests a dataset structure to fill a gap in the LPS.
    \item Execute: The code is run to generate the synthetic data.
    \item Repair: If the resulting dataset fails to land in the targeted region of the LPS, the LLM adjusts the parameters to "steer" the next generation more accurately.
\end{itemize}

This loop acts as the corrective mechanism for \textit{bridging the LLM and the manifold}, ensuring that the \textit{synthetic witness} datasets effectively populate the coordinates required for high-fidelity meta-learning. The relative failure of margin-based refinement confirms that the meta-learner requires a complete \textit{functional topography}, i.e., a map of the whole space, rather than just a refined view of the boundaries.

%\rev{This leads to a refinement of the $\epsilon$-cover condition. While a uniform $\epsilon$-cover is necessary to establish the \textit{global topology} of the manifold, the 'resolution' of the decision boundaries determines the final selection accuracy. In a low-data regime, margin-based sampling fails because it attempts to resolve boundaries on a manifold that has not yet been globally defined. Once the global 'scaffold' is established via uniform sampling, the marginal utility of additional uniform points diminishes, and the focus must shift to \textit{boundary sharpening}. The geometric limit of meta-learning generalisation is thus likely achieved through a curriculum that sequences these two objectives.}

%\rev{This synthesis formalises the relationship between coverage and refinement as a \textit{Geometric Curriculum}. If $M$ represents the performance manifold and $C$ represents the current synthetic cover, the optimal acquisition strategy $\mathcal{A}$ is governed by the state of manifold maturity:
%\[
%\mathcal{A} = \left\{ 
%\begin{array}{ll}
%\text{Uniform } (\epsilon\text{-cover}) & \text{if } \text{dist}(M, C) > \tau \\
%\text{Margin-based (Refinement)} & \text{if } \text{dist}(M, C) \leq \tau
%\end{array} 
%\right.
%\]
%where $\tau$ represents the threshold at which global topological stability is achieved.}

%\rev{This explains the early-stage failure of margin-based sampling observed in Section~\ref{sec:ablation_learning_curves}, i.e., that the strategy attempts to resolve boundaries on a manifold that has not yet reached topological stability.}

%% file: sections/conclusion.tex
In this work, we began with an investigation into mitigating the scarcity of meta-learning data through LLM-driven synthesis. However, our findings have evolved into a broader thesis regarding the nature of algorithm selection. More specifically, we notice that while individual dataset complexity matters, \textbf{the global coverage of the performance manifold is the primary driver of meta-learner generalisation}.

We demonstrated that while LLMs generate structurally simpler datasets, these datasets provide a critical "scaffold" for the meta-learner by acting as a representative sample of the functional landscape. Our comparison of sampling strategies revealed a clear hierarchy: \textbf{Uniform Performance-Space Sampling} consistently outperforms margin-based approaches.This suggests that for algorithm selection, the meta-learner requires a \textit{holistic functional topography}. Achieving an empirical $\epsilon$-cover of the performance manifold is the primary prerequisite for generalisation, as it allows the meta-learner to anchor its structural scaffolding before attempting local boundary refinement. By prioritising this "covering" over high-density refinement near decision boundaries, we achieved a $100.41\%$ improvement in subset accuracy for multi-label selection.

\subsection{Future Work}

The transition from data generation to performance-space mapping opens several avenues for future inquiry:

\begin{itemize}
    %\item \textbf{The Hybrid-Switch Implementation}: Based on the non-monotonic trajectories observed in Section 5, we propose the development of an \textit{Adaptive Switch Mechanism}. Instead of selecting a single strategy a priori, the synthesis engine should employ an \textit{Adaptive Switch Mechanism}. This engine monitors the stability of the structural embedding; upon identifying the plateau in pooled out-of-fold $R^2$ improvement under uniform sampling, the generator pivots to \textit{margin-based targeting}. This dual-phase approach leverages uniform sampling for global anchoring and margin-based sampling for \textit{high-resolution boundary definition}. This would allow the pipeline to benefit from both the global anchoring of uniform coverage and the high-resolution boundary definition of margin-based sampling, potentially bypassing the subset accuracy decline observed at high volumes.

    %Specifically, this switch represents the operationalisation of the 'Maturity Threshold'. By transitioning from expansion (minimising $\epsilon$) to refinement (maximising boundary precision) only after the scaffold is stable, the pipeline can achieve the geometric limit of meta-learning generalisation without the 'cold-start' penalties identified in our benchmarks.

    \item \textbf{Expanding the Generative Vocabulary}. To overcome the identified structural limits where the generator favours additive mixtures (Table 4), we propose a \textit{Prompt Chaining} architecture. Instead of monolithic dataset generation, the process would be decomposed into a three-stage sequence:
    
    \begin{enumerate}
        \item \textbf{Logic Selection}: Forcing the inclusion of non-linear operators (e.g., \texttt{np.where} for branching or \texttt{np.linalg} for spectral transforms).
        \item \textbf{Implementation}: Drafting the mechanism (script) based on the selected logic.
        \item \textbf{Validation}: Executing the code to ensure the resulting performance profile aligns with the target cell in the $\epsilon$-cover.
    \end{enumerate}
    
    This "\textit{Chain-of-Synthesis}" architecture, which incorporates logic selection, implementation, and iterative validation, ensures the generator traverses the full structural complexity of the algorithm selection space. This prevents the generative process from collapsing into a narrow subspace of additive Gaussian mixtures and forces the inclusion of diverse, non-linear operators.\\
    
    \item \textbf{Active Performance Mapping}: the success of uniform performance-space sampling suggests a move toward active performance mapping. In this framework, the LLM is not a static data source but an active query participant within an active learning loop. By quantifying the meta-learner’s uncertainty (e.g., through committee-based disagreement or high Hamming loss variance) across the performance manifold, we can direct the LLM to synthesise datasets specifically in those "low-confidence" regions. This transforms the LLM from a broad-spectrum augmentor into a precision tool for iteratively closing the $\epsilon$-cover gap.\\
    
    \item \textbf{Cross-Domain Generalisation}: The \textit{uniform-is-superior} hypothesis should be tested in other paradigms, such as classification and time-series forecasting, to determine if global topological coverage is a universal requirement for effective meta-learning.\\
    
    \item \textbf{Domain-Specific Narrative Injection}: Investigating whether grounding synthetic datasets in specific domain narratives (e.g., biological or financial constraints) changes the meta-learner’s ability to generalise to those specific fields, even if the underlying mathematical structure remains the same.\\
\end{itemize}

A natural extension is a wider sweep of grid granularity beyond the two grids used here. With the uniform-augmentation benefit already established, varying grid resolution more finely would test whether the regression-vs-classification split observed between the 5x5 and 7x7 grids is monotonic in grid resolution or reverses. A finer sweep would also identify where, if anywhere, a single granularity dominates across the regression and multi-label formulations simultaneously.

However, our immediate focus in the near future will be to work on a \textbf{theoretical formalisation of $\epsilon$-covers}. Our most immediate next steps will be the development a formal theoretical framework to quantify the density of the synthetic meta-dataset as an $\epsilon$-cover of the performance manifold. This would allow us to determine the minimum number of synthetic datasets required to guarantee a specific bound on meta-learning error for unseen real-world datasets.

%% file: sections/appendix.tex
\subsection{UCI regression datasets}
Table~\ref{tab:regression_datasets} lists the 42 UCI regression datasets used to construct the real-task component of our meta-dataset, together with their numbers of features and instances. Each dataset was cleaned and preprocessed with a uniform pipeline before being used to create the meta-dataset. Rows containing missing values were dropped. Categorical columns were one-hot encoded with \texttt{drop="first"} to avoid collinearity from the redundant reference category. Uninformative columns---such as date and timestamp fields, row indices, and identifier columns that carry no predictive signal---were removed. The feature and instance counts reported in Table~\ref{tab:regression_datasets} reflect the datasets after this preprocessing.

\begin{table}[ht]
\centering
\caption{The 42 UCI regression datasets used in the real-task meta-dataset. ``\# Features'' denotes the number of predictive columns (all columns except the regression target), and ``\# Instances'' denotes the number of rows in the CSV file excluding the header.}
\label{tab:regression_datasets}
\small
\setlength{\tabcolsep}{5pt}
\renewcommand{\arraystretch}{1.1}
\begin{minipage}[t]{0.49\linewidth}
\centering
\begin{tabular}{@{}p{4.2cm}rr@{}}
\toprule
\textbf{Dataset} & \textbf{\# Features} & \textbf{\# Instances} \\
\midrule
Air Quality (CO)                & 8   & 827  \\
Air Quality (NMHC)              & 8   & 827  \\
Air Quality (NO$_2$)            & 8   & 827  \\
Air Quality (NO$_x$)            & 8   & 827  \\
Air Quality (O$_3$)             & 8   & 827  \\
Airfoil Self-Noise              & 5   & 1503 \\
Auto MPG                        & 22  & 392  \\
Automobile                      & 58  & 159  \\
Bike Sharing (Daily)            & 29  & 731  \\
Combined Cycle Power Plant      & 4   & 9568 \\
Communities and Crime           & 124 & 319  \\
Communities and Crime (Unnorm.) & 127 & 319  \\
Computer Hardware               & 6   & 209  \\
Concrete Compressive Strength   & 8   & 1030 \\
Concrete Slump (Flow)           & 7   & 103  \\
Concrete Slump (Slump)          & 7   & 103  \\
Concrete Slump (Strength)       & 7   & 103  \\
CSM (Movies)                    & 24  & 187  \\
Daily Demand Forecasting Orders & 18  & 60   \\
Energy Efficiency (Cooling)     & 8   & 768  \\
Energy Efficiency (Heating)     & 8   & 768  \\
\bottomrule
\end{tabular}
\end{minipage}%
\hfill
\begin{minipage}[t]{0.49\linewidth}
\centering
\begin{tabular}{@{}p{4.2cm}rr@{}}
\toprule
\textbf{Dataset} & \textbf{\# Features} & \textbf{\# Instances} \\
\midrule
Facebook Metrics                              & 56  & 495  \\
Geographical Origin of Music (Lat.)           & 116 & 1059 \\
Geographical Origin of Music (Lat., Default)  & 68  & 1059 \\
Geographical Origin of Music (Long.)          & 116 & 1059 \\
Geographical Origin of Music (Long., Default) & 68  & 1059 \\
GPS Trajectories                              & 10  & 163  \\
Istanbul Stock Exchange                       & 7   & 536  \\
Parkinson Speech                              & 26  & 1040 \\
Parkinson Telemonitoring (Motor UPDRS)        & 19  & 5875 \\
Parkinson Telemonitoring (Total UPDRS)        & 19  & 5875 \\
Servo                                         & 10  & 167  \\
SML2010                                       & 29  & 4137 \\
Stock Portfolio (Abs.\ Win Rate)              & 6   & 315  \\
Stock Portfolio (Annual Return)               & 6   & 315  \\
Stock Portfolio (Excess Return)               & 6   & 315  \\
Stock Portfolio (Rel.\ Win Rate)              & 6   & 315  \\
Stock Portfolio (Systematic Risk)             & 6   & 315  \\
Stock Portfolio (Total Risk)                  & 6   & 315  \\
Student Performance (Math)                    & 75  & 395  \\
Student Performance (Portuguese)              & 75  & 649  \\
Yacht Hydrodynamics                           & 6   & 308  \\
\bottomrule
\end{tabular}
\end{minipage}
\end{table}

\subsection{LLM Prompts}
\label{sec:llm_prompts}

This section documents the prompts supplied to the large language model used by our synthetic dataset generator (Section~\ref{sec:llm_generator}). For reproducibility, we reproduce each prompt verbatim as stored in the project repository. Placeholders of the form \texttt{\{target\_description\}} and \texttt{\{achieved\_description\}} are runtime-substituted with the natural-language specification of the current target cell in the 2D performance space and, where applicable, the observed outcome of the previous generation attempt.

\subsubsection{System Prompt}

Sent as the \texttt{system} role for every call. Establishes the task, the 2D performance space, the evaluation tool available inside the generated code, code requirements, and the strict JSON output schema.

\newpage

\begin{lstlisting}[style=prompt]
You are an expert ML engineer designing synthetic regression dataset generators.

Your task is to produce Python code that defines:
    generate(seed: int) -> tuple[np.ndarray, np.ndarray]

The returned dataset (X, y) should be intentionally designed so that, when externally evaluated, it lands inside a specified target box in a 2D performance space.

The 2D performance space is defined as:
- x-axis: mean cross-validated R^2 of KNN regressor
- y-axis: mean cross-validated R^2 of Linear Regression

You will be given:
- target x-range
- target y-range
- target center
- a natural-language description of the target cell

Available tool inside the generated code:
- evaluate(X, y) -> {"x_score": float, "y_score": float}

Use evaluate(X, y) inside generate(seed) if helpful for search, tuning, or acceptance checking.

Requirements for the generated Python code:
- Define a function named generate(seed: int)
- Return a tuple (X, y)
- X must be a 2D NumPy array
- y must be a 1D NumPy array
- Deterministic given seed
- Self-contained
- Use NumPy only (already loaded)
- Do not use imports
- Do not read or write files
- Do not depend on any external libraries
- Keep the implementation concise but clear
- Add brief comments only where useful

Design objective:
- Use dataset mechanisms that deliberately control relative performance of KNN and Linear Regression
- Prefer mechanisms that are interpretable and adjustable
- If useful, construct candidate datasets and refine them using evaluate(X, y)
- Keep any search lightweight and deterministic
- The final returned dataset should be the best candidate found within the generate(seed) call

Output format:
Return exactly one valid JSON object and nothing else.

The JSON object must match this structure exactly:
- "mechanism_brief": string
- "python_code": string
- "expected_x_behavior": string
- "expected_y_behavior": string

Field requirements:
- "mechanism_brief": short explanation of the intended generation mechanism
- "python_code": Python code defining generate(seed: int) -> tuple[np.ndarray, np.ndarray]
- "expected_x_behavior": what behavior is expected along the x-axis metric
- "expected_y_behavior": what behavior is expected along the y-axis metric

Do not use XML-like tags.
Do not include markdown fences.
Do not include commentary before or after the JSON object.
Do not include additional keys.
\end{lstlisting}

\subsubsection{Initial Generation Prompt}

Sent as the first \texttt{user} message when requesting a generator for a new target cell.

\begin{lstlisting}[style=prompt]
Target specification:
<target_spec>
{target_description}
</target_spec>

Generate a synthetic regression dataset generator intended to place the dataset inside the target box.

Important:
- Use the target region as the primary objective
- Use the target center as the directional guide when choosing or tuning the mechanism
- Use your knowledge of how KNN and Linear Regression behave under different data-generating mechanisms
- You may use evaluate(X, y) inside generate(seed) to search over candidate constructions
- Prefer robust mechanisms over brittle one-off tricks

A good answer:
- proposes a plausible mechanism whose geometry matches the desired relative strengths of KNN and Linear Regression
- writes valid deterministic Python
- explains the expected x-axis behavior in "expected_x_behavior"
- explains the expected y-axis behavior in "expected_y_behavior"

Return exactly one valid JSON object matching the required schema, using the keys:
- "mechanism_brief"
- "python_code"
- "expected_x_behavior"
- "expected_y_behavior"

Return only JSON and no surrounding text.
\end{lstlisting}

\newpage

\subsubsection{Repair Prompt}

Sent as a follow-up \texttt{user} message when the generator produced by the previous attempt landed outside the target box. The achieved outcome is included so the model can reason about the required direction of adjustment.

\begin{lstlisting}[style=prompt]
Previous attempt missed the target region.

Target specification:
<target_spec>
{target_description}
</target_spec>

Observed outcome:
<observed_result>
{achieved_description}
</observed_result>

Revise the generator so the next attempt is more likely to land inside the target box.

Repair goal:
- determine whether x_score should increase or decrease
- determine whether y_score should increase or decrease
- adjust the mechanism accordingly
- make the smallest effective change when possible
- replace the mechanism entirely if it is poorly matched to the target
- preserve all code constraints
- return the best candidate found by the generated function

Return exactly one valid JSON object matching the required schema, using the keys:
- "mechanism_brief"
- "python_code"
- "expected_x_behavior"
- "expected_y_behavior"

Return only JSON and no surrounding text.
\end{lstlisting}

\lstdefinestyle{codepy}{
  basicstyle=\ttfamily\footnotesize,
  breaklines=true,
  breakatwhitespace=false,
  columns=fullflexible,
  keepspaces=true,
  frame=single,
  framesep=4pt,
  xleftmargin=6pt,
  xrightmargin=6pt,
  showstringspaces=false,
  upquote=true,
  language=Python,
  keywordstyle=\bfseries,
  commentstyle=\itshape\color{posgreen},
  stringstyle=\color{posbrown},
}

\subsection{Synthetic Dataset Generation: Analysis and Examples}
\label{sec:synthetic_analysis}

This appendix documents the synthetic datasets accepted by the LLM-based generator described in Section~\ref{sec:llm_generator}. The generator targets a $7 \times 7$ grid of cells in the 2D performance space (cross-validated $R^2$ of KNN on the $x$-axis, cross-validated $R^2$ of Linear Regression on the $y$-axis); for each cell it proposes a sequence of candidate generator functions until up to ten datasets land inside the target box. Accepted datasets are stored in a two-level directory hierarchy: the outer level is the cell (\texttt{cell\_ii\_jj}, with \texttt{ii} indexing the LR bin and \texttt{jj} indexing the KNN bin), and the inner level is the accepted attempt, each carrying the generated CSV, a JSON metadata file (including the LLM's natural-language mechanism brief), and the Python generator function. The subsections below summarise the contents of the 484 accepted datasets, extract the themes and taxonomy of the data-generating mechanisms the LLM invents, compare the synthetic meta-feature distributions to the real-task distributions, and reproduce three representative generator functions.

\subsubsection{Accepted datasets: coverage and size}

Across the $7 \times 7 = 49$ target cells the generator produced 484 accepted datasets. Forty-seven cells were filled to the configured quota of ten datasets each; two cells (\texttt{cell\_03\_06} and \texttt{cell\_06\_00}) were declared \emph{exhausted} after 84 attempts each yielded only seven accepted datasets. Per-dataset size summaries are given in Table~\ref{tab:synth_size_summary}.

\begin{table}[ht]
\centering
\caption{Summary of dataset sizes across the 484 accepted synthetic datasets. ``\#~Features'' counts predictive columns only. All datasets are regression tasks with a single continuous target.}
\label{tab:synth_size_summary}
\small
\begin{tabular}{@{}lrrr@{}}
\toprule
\textbf{Attribute} & \textbf{Min} & \textbf{Mean} & \textbf{Max} \\
\midrule
\# Instances (rows) & 90 & 213.8 & 520 \\
\# Features         & 1  & 19.1  & 342 \\
\bottomrule
\end{tabular}
\end{table}

The feature-count distribution is strongly right-skewed and cell-dependent: cells targeting the KNN-favouring corner of the performance space use very few features (one to three informative features and essentially no distractors), whereas cells targeting the LR-favouring corner rely on 20--70 (occasionally over 300) irrelevant Gaussian dimensions to suppress KNN through distance concentration. Row counts cluster near 200 because the generator's grid-search defaults select moderate sample sizes that give stable cross-validation scores without over-extending per-attempt runtime.

\subsubsection{Mechanism-brief themes}

The keyword-scan findings in this paragraph are interpreted in Section~\ref{sec:canonical_characterisation} as evidence of the structural limits of LLM data synthesis; the data below provides the supporting evidence. Every accepted JSON file carries a one-paragraph \emph{mechanism brief} written by the LLM summarising the intended data-generating process.
A keyword scan over the 484 briefs reveals a narrow stylistic repertoire. The phrases ``irrelevant / nuisance / distractor features'' appear in roughly \num{60}\% of briefs, ``local / neighbourhood structure'' in roughly \num{49}\%, and ``additive / mixture / linear-plus-smooth'' phrasing in roughly \num{28}\%. Conversely, \emph{zero} briefs dress the data in any domain-specific narrative: the LLM never invokes a medical, financial, physical, or other applied-context story under the current prompt. Tree-structured, polynomial-interaction, and probabilistic-draw target constructions are also almost entirely absent. In short, the LLM defaults to a small family of statistics-textbook mechanisms parameterised by two dials: (i) how curved or non-monotone the target is (which suppresses Linear Regression), and (ii) how many irrelevant feature dimensions accompany the signal (which suppresses KNN). Briefs written during repair attempts are easy to identify because they start with phrases such as ``Keep the same \ldots{} but slightly \ldots,'' reflecting a strong preference for minimal directional adjustment over wholesale mechanism replacement.

\subsubsection{Taxonomy of generated mechanisms}

Classifying each brief by its primary mechanism family (with a priority rule so that more specific mechanisms win) produces the eleven categories shown in Table~\ref{tab:synth_taxonomy}. Counts sum to 484; percentages may not add to 100 because of rounding. The right-most column reports the centroid of each category's cell locations in grid coordinates $(\textsc{knn\_bin}, \textsc{lr\_bin})$, where each bin index ranges over $0 \le \textsc{bin} \le 6$.

\begin{table}[ht]
\centering
\caption{Taxonomy of data-generating mechanisms inferred inductively from the 484 accepted mechanism briefs. Each brief is assigned to a single primary category by priority rule.}
\label{tab:synth_taxonomy}
\footnotesize
\setlength{\tabcolsep}{4pt}
\renewcommand{\arraystretch}{1.2}
\begin{tabular}{@{}p{2.6cm}p{4.6cm}p{4.2cm}rrp{1.5cm}@{}}
\toprule
\textbf{Category} & \textbf{Core idea} & \textbf{Representative paraphrase} & \textbf{\#} & \textbf{\%} & \textbf{Cell centroid} \\
\midrule
M1 Additive mixture & Linear backbone plus smooth nonlinear term. & ``Mix a linear signal with a smooth local sinusoid\ldots'' & 105 & 21.7 & (3.6, 3.2) \\
M2 Smooth 1-D nonlinear & Single curved feature (sinusoid, cosine, quadratic, U-shape). & ``y is mainly a sinusoid of one feature\ldots'' & 62 & 12.8 & (4.1, 1.7) \\
M3 Weak linear + nuisance-dominated & One weak linear feature drowned in 20--70 Gaussian distractors. & ``Many irrelevant features plus one weak linear signal\ldots'' & 54 & 11.2 & (2.1, 4.0) \\
M4 Radial / 2-D localised bump & Gaussian/radial bump over two informative features. & ``Radial bump that KNN can capture well\ldots'' & 53 & 11.0 & (3.9, 2.5) \\
M5 Near-zero / pure noise & Target nearly independent of $X$. & ``Pure-noise target largely independent of features\ldots'' & 52 & 10.7 & (1.8, 1.6) \\
M6 Piecewise / step / V-shape / checkerboard & Discontinuous or kinked target (hinge, absolute value, sign quadrant). & ``Checkerboard signal on two latent dimensions\ldots'' & 23 & 4.8 & (3.6, 2.0) \\
M7 Predominantly linear (low-dim) & Mostly-linear signal with few features and few distractors. & ``Mostly-linear signal with several irrelevant features\ldots'' & 20 & 4.1 & (3.5, 3.6) \\
M8 Heteroscedastic / adversarial noise & Feature-conditioned or heavy-tailed noise to selectively degrade KNN. & ``Heteroscedastic noise driven by an irrelevant feature\ldots'' & 17 & 3.5 & (2.7, 4.8) \\
M9 Label corruption & Duplicated rows with conflicting labels. & ``Duplicated feature rows with randomly assigned responses\ldots'' & 15 & 3.1 & (2.3, 3.3) \\
M10 Latent factor + redundant copies & Scalar latent $z$ observed through several noisy copies. & ``One-factor latent regression with noisy copies\ldots'' & 7 & 1.4 & (1.6, 3.0) \\
M11 Other / unclassifiable & Briefs too short or too hybrid for a single label. & --- & 76 & 15.7 & scattered \\
\bottomrule
\end{tabular}
\end{table}

The mechanism patterns analysed in this paragraph are interpreted in Section~\ref{sec:canonical_characterisation} as evidence of the structural limits of LLM data synthesis; the analysis below provides the supporting data. Three qualitative patterns emerge. First, the mechanism family correlates cleanly with cell location: families M2, M4, and M6 (strong nonlinearity) cluster in high-KNN cells, M3 and M8 (nuisance-heavy) cluster in high-LR cells, and M1 (additive mixture) dominates the diagonal where both models get partial signal. Second, the LLM rarely replaces a mechanism once it produces something close to the target --- repair briefs preserve the mechanism family in the overwhelming majority of cases and tune only noise scale, signal amplitude, and irrelevant-dimension count. Third, the secondary axes (feature-construction strategy, noise model, complexity) are dominated by a single default configuration: additive homoscedastic Gaussian noise, independent Gaussian features, no correlated feature blocks, no categorical variables, and no probabilistic targets. The LLM's ``creativity'' under the current prompt is therefore narrow: it explores a small manifold of mechanisms by moving along two well-defined dials rather than by inventing qualitatively new families.

\subsubsection{Meta-feature comparison: real vs.\ synthetic}

\begin{figure}[ht]
\centering
\includegraphics[width=\linewidth]{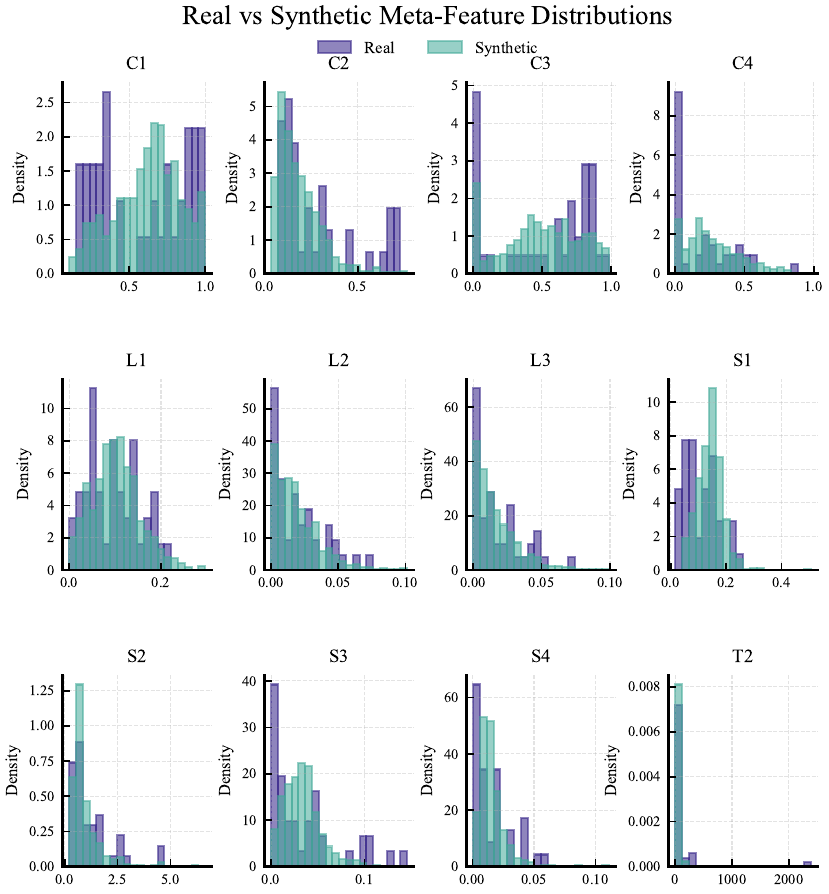}
\caption{Histograms of selected meta-features computed on the 42 real UCI regression datasets and on the 484 accepted synthetic datasets. The synthetic datasets cover a comparable meta-feature range but with visibly different marginal shapes, reflecting the fact that the generator is guided by target performance-space cells rather than by meta-feature matching.}
\label{fig:real_vs_synthetic_hist}
\end{figure}

Figure~\ref{fig:real_vs_synthetic_hist} overlays the distributions of selected meta-features computed on the 42 real UCI regression datasets against the corresponding distributions over the 484 accepted synthetic datasets. The comparison is useful for reproducibility because the generator is explicitly \emph{not} optimised to match the real-task meta-feature distribution --- it is optimised to cover the 2D performance space.

\subsubsection{Generator run summary}

Aggregate statistics of the single run are listed in Table~\ref{tab:synth_run_summary}. The run made 1{,}729 attempts to fill 49 cells, yielding 484 accepted datasets at an overall hit rate of \num{28.0}\%; per-cell hit rates ranged from \num{8.3}\% (the two exhausted cells) to \num{100}\% (\texttt{cell\_02\_02} and \texttt{cell\_03\_03}, where every attempt landed in-box). Total token usage across the run was \num{6.66} million tokens, or roughly \num{3858} tokens per attempt.

\begin{table}[ht]
\centering
\caption{Aggregate statistics for the synthetic-data generation run. Means and totals are over all 1{,}729 attempts; cell-level fill counts are over the 49 target cells.}
\label{tab:synth_run_summary}
\small
\begin{tabular}{@{}lr@{}}
\toprule
\textbf{Metric} & \textbf{Value} \\
\midrule
Run ID                                        & \texttt{pilot\_7x7\_v1\_\_20260403\_085503} \\
Target cells ($7 \times 7$ grid)              & 49 \\
Total attempts                                & 1{,}729 \\
Accepted datasets                             & 484 \\
Overall hit rate                              & 0.280 \\
Mean attempts per accepted dataset            & 3.52 \\
Cells fully filled (10/10)                    & 47 \\
Exhausted cells (quota not met)               & 2 \\
Mean LLM runtime per attempt                  & \SI{13.56}{\second} \\
Mean code-execution runtime per attempt       & \SI{26.69}{\second} \\
Mean evaluation runtime per attempt           & \SI{0.04}{\second} \\
Mean total runtime per attempt                & \SI{40.28}{\second} \\
Total input tokens                            & 5{,}102{,}088 \\
Total output tokens                           & 1{,}560{,}892 \\
Total tokens                                  & 6{,}662{,}980 \\
Mean input tokens per attempt                 & 2{,}954 \\
Mean output tokens per attempt                & 904 \\
\bottomrule
\end{tabular}
\end{table}

The logs do not record a per-attempt rejection-reason field; a rejection is simply any attempt whose achieved $(\hat{R}^2_{\mathrm{KNN}}, \hat{R}^2_{\mathrm{LR}})$ lies outside the target cell's box. The two exhausted cells both sit in extreme corners of the performance grid where the geometric constraints on the data-generating mechanism are hardest to satisfy simultaneously.

\subsubsection{Representative generated algorithms}

To illustrate the taxonomy concretely we reproduce three accepted generator functions drawn from three different taxonomy categories (M1, M3, and M6) and three different regions of the performance grid. Each listing is reproduced verbatim; the implicit \texttt{np} reference and the injected \texttt{evaluate(X, y)} tool are as specified in the system prompt (Section~\ref{sec:llm_prompts}).

\paragraph{Sample 1 --- Additive mixture (M1).} Source dataset: \texttt{cell\_02\_03\_\_ds\_006} (first-attempt accept). Achieved scores $(\hat{R}^2_{\mathrm{KNN}}, \hat{R}^2_{\mathrm{LR}}) = (0.432, 0.401)$ on 220 rows $\times$ 6 features (2 informative features, 4 iid Gaussian distractors). Mechanism brief:
\begin{quote}
\itshape ``Moderately noisy additive signal mixing a global linear trend with a smooth local sinusoid in one feature plus several nuisance dimensions. The linear part keeps Linear Regression moderately predictive, while the sinusoid gives KNN a modest edge without making either model too strong.''
\end{quote}

\begin{lstlisting}[style=codepy, caption={M1 additive-mixture generator (\texttt{cell\_02\_03\_\_target\_006\_\_att\_001.py}).}, label={lst:synth_m1}]
def generate(seed: int):
    rng = np.random.RandomState(seed)

    def build(n, d_noise, lin, amp, noise, x_scale, extra_nl):
        x0 = rng.uniform(-x_scale, x_scale, n)
        x1 = rng.uniform(-x_scale, x_scale, n)
        z = rng.normal(0, 1, (n, d_noise))
        y = lin * x0 + amp * np.sin(2.2 * x1) + extra_nl * 0.25 * (x0 * x1) + noise * rng.normal(0, 1, n)
        X = np.column_stack([x0, x1, z])
        return X, y

    target_x = 0.50
    target_y = 0.357143
    best = None
    best_dist = 1e9

    # Small deterministic grid around an interpretable mechanism.
    for n in [180, 220, 260]:
        for d_noise in [4, 6, 8]:
            for lin in [1.0, 1.15, 1.3]:
                for amp in [0.6, 0.8, 1.0]:
                    for noise in [1.15, 1.35, 1.55]:
                        for extra_nl in [0.0, 0.6]:
                            X, y = build(n, d_noise, lin, amp, noise, 2.2, extra_nl)
                            try:
                                s = evaluate(X, y)
                                dx = s['x_score'] - target_x
                                dy = s['y_score'] - target_y
                                dist = dx * dx + dy * dy
                                in_box = (0.428571 <= s['x_score'] <= 0.571429 and
                                          0.285714 <= s['y_score'] <= 0.428571)
                                if in_box:
                                    dist *= 0.2
                                if dist < best_dist:
                                    best_dist = dist
                                    best = (X, y)
                            except Exception:
                                if best is None:
                                    best = (X, y)

    return best
\end{lstlisting}

The mechanism is realised literally: a linear term in \texttt{x0}, a sinusoidal term in \texttt{x1}, an optional weak interaction \texttt{x0*x1}, additive Gaussian noise, and a block of \texttt{d\_noise} iid distractor columns. The outer six-deep grid evaluates each candidate with the injected \texttt{evaluate(...)} tool and keeps the candidate whose $(x, y)$ score is closest to the cell centre, with a $5\times$ distance discount applied if the candidate lands strictly inside the target box --- a simple hand-rolled implementation of ``prefer in-box solutions'' that is mentioned in many briefs.

\paragraph{Sample 2 --- Weak linear + nuisance-dominated (M3).} Source dataset: \texttt{cell\_04\_00\_\_ds\_005} (first-attempt accept). Achieved scores $(\hat{R}^2_{\mathrm{KNN}}, \hat{R}^2_{\mathrm{LR}}) = (0.120, 0.617)$ on 140 rows $\times$ 41 features (1 informative + 40 iid Gaussian distractors) --- the canonical curse-of-dimensionality cell. Mechanism brief:
\begin{quote}
\itshape ``Use many irrelevant noise features plus one moderately predictive linear feature, then add substantial target noise. Linear regression can still recover the weak global linear signal, while KNN is hurt badly by distance concentration in the high-dimensional space.''
\end{quote}

\begin{lstlisting}[style=codepy, caption={M3 weak-linear-plus-nuisance generator (\texttt{cell\_04\_00\_\_target\_005\_\_att\_001.py}).}, label={lst:synth_m3}]
def generate(seed: int):
    rng = np.random.RandomState(seed)

    def make_data(n, d_noise, beta, noise):
        x0 = rng.normal(size=n)
        noise_feats = rng.normal(size=(n, d_noise))
        X = np.column_stack([x0, noise_feats])
        y = beta * x0 + rng.normal(scale=noise, size=n)
        return X, y

    target_x = 0.071429
    target_y = 0.642857
    best = None
    best_dist = 1e9

    # Search for: weak linear signal + many irrelevant dimensions.
    for n in [140, 180, 220, 260]:
        for d_noise in [20, 30, 40, 55, 70, 90]:
            for beta in [1.2, 1.5, 1.8, 2.1, 2.4, 2.8, 3.2]:
                for noise in [2.0, 2.3, 2.6, 2.9, 3.2, 3.6]:
                    X, y = make_data(n, d_noise, beta, noise)
                    try:
                        scores = evaluate(X, y)
                        xs = float(scores['x_score'])
                        ys = float(scores['y_score'])
                    except Exception:
                        xs = ys = -1e9

                    in_box = (0.0 <= xs <= 0.142857) and (0.571429 <= ys <= 0.714286)
                    dist = (xs - target_x) ** 2 + (ys - target_y) ** 2
                    if in_box:
                        return X, y
                    if dist < best_dist:
                        best_dist = dist
                        best = (X, y)

    return best
\end{lstlisting}

Here the mechanism is a single arithmetic line, \texttt{y = beta * x0 + noise}, and every feature other than \texttt{x0} is a pure Gaussian distractor. The search sweeps sample size, distractor count, signal strength, and noise scale. Critically, the target-box test triggers an immediate early return, which is a more aggressive acceptance policy than Sample 1's soft penalty. Together with Sample 1 this illustrates the range of search logic the LLM writes --- from simple early-exit to soft multi-objective penalties.

\paragraph{Sample 3 --- Piecewise / V-shape (M6).} Source dataset: \texttt{cell\_01\_06\_\_ds\_004} (accepted on the third attempt after two repairs). Achieved scores $(\hat{R}^2_{\mathrm{KNN}}, \hat{R}^2_{\mathrm{LR}}) = (0.930, 0.214)$ on 280 rows $\times$ 3 features (1 latent-carrying feature plus 2 distractors) --- the strongly KNN-favouring edge. Mechanism brief:
\begin{quote}
\itshape ``Raise KNN while lowering Linear Regression by using a strongly local but globally weak relationship: a dominant absolute-value response in one feature, plus only a very small linear component, low observation noise, and a couple of irrelevant features. KNN can fit the V-shape well, while Linear Regression only captures the tiny residual trend.''
\end{quote}

\newpage

\begin{lstlisting}[style=codepy, caption={M6 piecewise / V-shape generator (\texttt{cell\_01\_06\_\_target\_004\_\_att\_003.py}).}, label={lst:synth_m6}]
def generate(seed: int):
    rng = np.random.RandomState(seed)

    tx, ty = 0.928571, 0.214286
    xlo, xhi = 0.857143, 1.0
    ylo, yhi = 0.142857, 0.285714

    def build(n, a_abs, a_lin, noise, d_noise, x_noise):
        z = rng.uniform(-1.0, 1.0, size=n)
        X = rng.normal(0.0, 1.0, size=(n, d_noise + 1))
        X[:, 0] = z + x_noise * rng.normal(size=n)
        y = a_abs * (np.abs(z) - 0.5) + a_lin * z + noise * rng.normal(size=n)
        return X, y

    candidates = []
    for a_abs in [0.9, 1.1, 1.3, 1.5]:
        for a_lin in [0.18, 0.24, 0.30, 0.36]:
            for noise in [0.04, 0.06, 0.08]:
                for d_noise in [1, 2, 3]:
                    for x_noise in [0.0, 0.02, 0.04]:
                        X, y = build(280, a_abs, a_lin, noise, d_noise, x_noise)
                        sc = None
                        try:
                            sc = evaluate(X, y)
                        except Exception:
                            pass
                        candidates.append((X, y, sc))

    if candidates and candidates[0][2] is not None:
        best = None
        best_loss = 1e18
        for X, y, sc in candidates:
            xs = float(sc['x_score'])
            ys = float(sc['y_score'])
            in_box = (xlo <= xs <= xhi) and (ylo <= ys <= yhi)
            dx = xs - tx
            dy = ys - ty
            loss = (0.0 if in_box else 10.0) + dx * dx + 1.3 * dy * dy
            if loss < best_loss:
                best_loss = loss
                best = (X, y)
        return best

    return build(280, 1.2, 0.26, 0.06, 2, 0.02)
\end{lstlisting}

The construction hides a latent $z$ inside the first feature column (\texttt{X[:, 0] = z + x\_noise * rng.normal}) and drives the target through \texttt{abs(z) - 0.5}, a centred V-shape whose global linear correlation with $z$ is near zero. Linear Regression therefore picks up only the small residual \texttt{a\_lin * z} tilt, while KNN can exploit the one-dimensional V-geometry almost perfectly. The search collects all candidates first and then ranks them, with a heavier additive penalty on out-of-box candidates (the additive $10$) and an asymmetric weighting of $y$-error (the $1.3\times$ factor) --- an example of the LLM hand-tuning its ranking criterion to the shape of the current target cell.

\subsubsection{Overall observations on the generated code}

Across all 484 Python generators we observe a remarkably consistent coding style:

\begin{itemize}
\item \textbf{Deterministic random state.} All 484 generators use \texttt{np.random.RandomState(seed)} and none use \texttt{np.random.default\_rng}, consistent with the deterministic-output requirement in the system prompt.
\item \textbf{evaluate() is used universally.} Every generator calls \texttt{evaluate(X, y)} at least once; \num{88}\% wrap the call in \texttt{try / except}, a defensive pattern that appears in essentially every generator from the second attempt onward.
\item \textbf{Helper plus grid.} \num{72}\% of generators factor the data construction into a helper function (\texttt{build}, \texttt{make\_data}) and drive a parameter grid; \num{51}\% explicitly track a ``best'' candidate across the grid. All search grids are small, finite, and deterministic, as required by the prompt.
\item \textbf{Acceptance-policy variation.} Within the grid-search pattern we see three recurring acceptance styles: (i) early-return on the first in-box hit (Sample 2), (ii) soft distance-based ranking with a multiplicative bonus for in-box candidates (Sample 1), and (iii) collect-then-rank with an additive out-of-box penalty (Sample 3). This is the main locus of code-level variability.
\item \textbf{Narrow numerical vocabulary.} This finding is interpreted in Section~\ref{sec:canonical_characterisation} as evidence of the structural limits of LLM data synthesis. The dominant NumPy primitives are \texttt{np.sin} (\num{50}\%), \texttt{np.hstack} or \texttt{np.column\_stack} (\num{59}\%), \texttt{np.exp} (\num{22}\%), and \texttt{np.cos} (\num{12}\%). Operators that would support qualitatively richer mechanisms --- \texttt{np.sign}, \texttt{np.where}, \texttt{np.tanh}, \texttt{np.linalg.*} --- appear in under \num{2}\% of generators. No generator imports \texttt{scipy}, \texttt{sklearn}, or any other library, in line with the prompt's no-imports constraint.
\item \textbf{Code size.} Median generator length is 54 lines (range 35--103). Longer generators typically correspond to cells in the hardest corners of the performance grid where the LLM resorts to deeper nested grid searches rather than to more complex mechanisms.
\end{itemize}

No generator exhibits an obvious correctness bug in its data construction; the most common failure mode visible in the logs is geometric rather than syntactic --- the proposed mechanism simply does not hit the target cell, and the attempt is rejected. Because rejection reasons are not logged structurally, this analysis is necessarily confined to what is visible in the accepted outputs.